\documentclass[11pt]{article}

\usepackage[T1]{fontenc}
\usepackage{lmodern}
\usepackage{microtype}
\usepackage[a4paper,margin=1in]{geometry}
\usepackage{amsmath,amssymb}
\usepackage{graphicx}
\usepackage{subcaption}
\usepackage{booktabs}
\usepackage{array}
\usepackage{multirow}
\usepackage{algorithm}
\usepackage{algpseudocode}
\usepackage{xurl}
\usepackage[authoryear,round]{natbib}
\usepackage[colorlinks=true,linkcolor=blue,citecolor=blue,urlcolor=blue]{hyperref}

\title{Sparse Federated Echo State Networks for Time-Series Classification via Smooth-L1 Regularization}

\author{Gengcheng Lyu\thanks{Corresponding author: \href{mailto:gengcheng.lyu@thws.de}{gengcheng.lyu@thws.de}}
  \and Manuel R\"oder \and Frank-Michael Schleif\\[1ex]
  \small Faculty of Computer Science and Business Information Systems\\
  \small Technical University of Applied Sciences W\"urzburg-Schweinfurt, W\"urzburg, Germany}
\date{}

\begin{document}

\maketitle

\begin{abstract}
Echo State Networks (ESNs), as lightweight implementations of reservoir computing, provide efficient solutions for temporal data processing in resource-constrained environments. While standard ESNs offer computational efficiency, their dense readout layers inflate the stored model and the per-sample inference cost. This paper introduces a Smooth-L1 regularized ESN (SL1ESN) that promotes sparsity in the readout layer while preserving differentiability for gradient-based optimization. The method employs a twice-differentiable Smooth-L1 approximation compatible with Newton's method, which we exploit both for centralized training and for a second-order (Newton-step) federated aggregation scheme. Under a single fixed shrinkage threshold shared by every dataset and both settings, experiments on seven time-series classification benchmarks show that SL1ESN learns readouts that are $77$--$95\%$ sparse in the centralized setting and $75$--$95\%$ sparse in the federated one. On five of the seven benchmarks, the mean accuracy of the sparse readout over ten seeds differs from that of the stronger dense baseline by at most $0.7$ points in the centralized setting and $1.6$ points in the federated one, differences of the same order as the seed-to-seed variation, while on the remaining two it trails by $2.4$--$5.8$ points. The second-order federated update converges in only a few rounds. The resulting sparsity compresses the stored model, the per-sample inference cost, and the server-to-client broadcast by a comparable factor. These results demonstrate the practical value of sparse readouts for edge computing and IoT deployments.
\end{abstract}

\noindent\textbf{Keywords:} Echo State Networks, Reservoir Computing, Sparse Models, Smooth-L1 Regularization, Temporal Sequence Modeling, Federated Learning

\section{Introduction}\label{sec1}

Machine learning has become a cornerstone for extracting patterns from complex datasets. While a large share of applied machine learning operates on tabular, independently sampled records, an important class of problems is instead inherently sequential, in the sense that the measurements arrive as a time-ordered stream and the ordering itself carries the information. Sensor and IoT telemetry, physiological recordings such as ECG, industrial process traces, and financial series all fall into this class, and they are also precisely the settings addressed in this paper, where the data are generated continuously and often on distributed, resource-constrained devices. Recurrent Neural Networks (RNNs) have long been a natural choice for such temporal processing because their hidden-state dynamics allow them to capture dependencies across time. However, traditional RNNs are often difficult to train reliably on long sequences and can suffer from vanishing or exploding gradients, which hampers the learning of long-range dependencies~\citep{pascanu2013difficulty,zucchet2024rnn,tiezzi2025statespace}. Gated architectures such as Long Short-Term Memory (LSTM)~\citep{hochreiter1997long} and Gated Recurrent Units (GRU)~\citep{chung2014empirical} alleviate some of these optimization challenges, and more recent designs continue this line. The xLSTM architecture~\citep{beck2024xlstm} augments the LSTM cell with exponential gating and matrix memory to recover much of the scaling behavior of Transformers while retaining a recurrent, constant-memory inference path, and deep state-space models pursue a similar goal from the linear-dynamics side~\citep{tiezzi2025statespace}. All of these nevertheless remain fully trained sequence models. They introduce a large number of trainable parameters and require backpropagation through time, with the associated computational and memory demands. These costs can become prohibitive in resource-constrained or real-time deployment scenarios, where compact and efficient temporal models remain especially desirable~\citep{lu2025survey,cordova2025edge}. Echo State Networks (ESNs) offer a fundamentally different response to the same difficulty. They are not a reaction to the recent architectures above but a parallel development. Introduced by~\cite{jaeger2001echo} at essentially the same time as gated recurrent models were becoming established, they sidestep the training difficulties of RNNs rather than engineering around them, and they remain a highly efficient class of recurrent models within the broader paradigm of Reservoir Computing (RC)~\citep{lukovsevivcius2009reservoir}. The core idea of RC is to separate the system into a fixed, randomly initialized recurrent layer, known as the reservoir, and a trainable output layer, called the readout. The reservoir acts as a high-dimensional dynamic memory that nonlinearly transforms the input signal into a rich state space. In standard ESNs, only the readout-layer weights are trained, often via ridge regression. By decoupling recurrent state generation from output-layer optimization, this design avoids direct training of recurrent connections and thereby reduces many of the optimization difficulties commonly associated with conventional RNNs. As a result, ESN training is typically much faster and often more stable than that of fully trained recurrent architectures~\citep{liu2026reservoir,matzner2025lcesn,yadav2024gestureesn}. Despite these significant advantages, standard ESN implementations employ a dense readout layer, meaning every reservoir neuron is connected to every output unit. While this ensures full utilization of the reservoir's representational power, it carries several drawbacks. First, dense readouts are prone to overfitting, since every reservoir neuron, including redundant ones, contributes to the output. Second, dense readouts increase the model's memory footprint and computational cost during inference. These limitations are particularly relevant in resource-constrained environments such as edge devices and Internet of Things (IoT) sensors, and in distributed or federated computing scenarios where communication bandwidth is often a bottleneck~\citep{dritsas2025fliot,akhtarshenas2024flsurvey,DBLP:journals/ijon/PolatoHRS26}.

To address these issues, sparsity-promoting methods in neural networks have been extensively studied~\citep{10328608,ijcai2020p735,DBLP:conf/esann/VillmannSH10}. In particular, L1 regularization~\citep{tibshirani1996regression} is well suited to promoting sparse solutions by driving irrelevant weights exactly to zero. Applying L1 regularization to the readout layer of an ESN yields a sparse model that is more memory-efficient and better suited for efficient inference. Sparsity is often also motivated by interpretability; because the reservoir features are strongly correlated, however, many different supports realize an equally sparse and accurate readout, so throughout this paper we claim compression rather than feature selection (Section~\ref{sec:limitations}). However, the L1 penalty is non-differentiable at zero, which complicates gradient-based optimization. While specialized algorithms such as Least Angle Regression (LARS)~\citep{efron2004least} and Coordinate Descent (CD)~\citep{Wright2015} solve the LASSO reliably in a centralized setting, they fit poorly into distributed optimization frameworks such as FedAvg~\citep{mcmahan2017}, a Federated Learning (FL) approach whose aggregation step is an average of locally computed, smooth parameter or gradient updates. The mismatch is structural rather than a matter of implementation. First, both algorithms are path- or coordinate-sequential, since LARS builds the solution by adding one variable to the active set at a time and CD updates one coordinate while holding the others fixed, so neither exposes a single descent direction over all coordinates that a client could compute locally and a server could average. Second, their state is a discrete active set rather than a vector in a linear space; clients trained on different local data return different supports, and the average of models with disjoint supports is neither sparse nor the solution of any of the local problems, so the aggregation step loses the very property the penalty was introduced to obtain. Third, and specific to our setting, neither method produces the curvature information that a second-order server update requires. Because the L1 term is non-differentiable at zero, no Hessian of the full objective exists that clients could compute and sum.

To overcome this limitation, we propose a novel ESN variant employing Smooth-L1 regularization\footnote{This article is an extended version of our earlier conference paper~\citep{DBLP:conf/ideal/LyuRS25}, with substantial additional methodological detail and expanded experimental results, including the federated learning setting.}. Smooth-L1 is a twice-differentiable approximation of the L1 norm that retains sparsity-inducing properties while remaining compatible with smooth, gradient-based optimization~\citep{schmidt2007fast}. This differentiability is crucial for efficient training via Newton's method~\citep{boyd2004convex} and for seamless integration into FL pipelines.

The main contributions of this paper are:
\begin{itemize}
\item \textbf{SL1ESN Architecture}: A novel ESN readout trained with
  a twice-differentiable Smooth-L1 penalty via damped Newton steps under a
  continuation schedule, which yields a sparse readout while keeping the
  training objective smooth and amenable to second-order optimization.

\item \textbf{Federated Second-Order Extension}: A distributed SL1ESN
  formulation, FedSL1ESN, in which clients return local gradients and
  Gauss-Newton Hessians and the server performs a damped Newton step with line
  search and shrinkage. The smoothness of the penalty is what makes this
  second-order federated update possible; it converges within a few
  communication rounds and yields a sparse global readout that reduces model
  storage, inference cost, and the server-to-client broadcast.

\item \textbf{Comprehensive Evaluation}: Extensive experiments on seven
  time-series classification tasks from the UCR and UCI archives,
  complemented by ablation, sensitivity, and client-count studies. The
  results demonstrate a favorable sparsity--accuracy trade-off, with readout sparsity of $77$--$95\%$ centrally and $75$--$95\%$ in the federated setting at accuracy close to the dense baselines, obtained under a single fixed shrinkage threshold rather than per-dataset tuning.
\end{itemize}
The remainder of the paper is organized as follows. Section~\ref{sec2} reviews related work. Section~\ref{sec3} introduces the ESN background and the proposed Smooth-L1 regularization. Section~\ref{sec4} presents the centralized and distributed training algorithms. Section~\ref{sec5} describes the experimental setup and results. Section~\ref{sec6} discusses the findings, and Section~\ref{sec7} concludes the paper.

\begin{figure*}[htbp]
  \centering
  \includegraphics[width=0.85\textwidth]{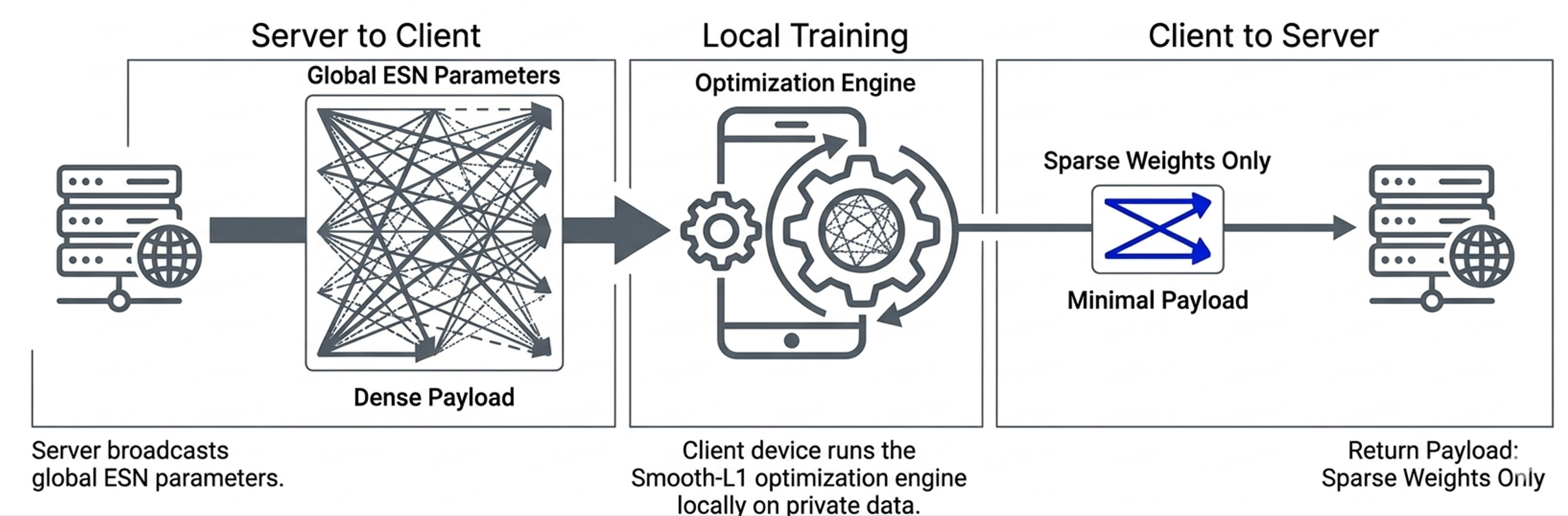}
  \caption{FedSL1ESN training pipeline. The server broadcasts
    the fixed global reservoir once and a sparse global readout each round
    (CSR, minimal payload); clients evaluate the local gradient and
    Gauss-Newton Hessian of their Smooth-L1 loss on private data and return
    this curvature information for the server's Newton-step aggregation.}
  \label{fig:sparse_readout}
\end{figure*}

\section{Related Work}\label{sec2}

\subsection{Reservoir Computing and Echo State Networks}

Echo State Networks are a foundational model within Reservoir Computing (RC). The ESN architecture consists of three components: an input layer, a fixed randomly initialized recurrent reservoir, and a trainable readout layer. The input is projected into the reservoir space via an input weight matrix $\mathbf{W}_{\text{in}}$, and the reservoir's internal dynamics evolve according to a fixed weight matrix $\mathbf{W}_{\text{res}}$, ensuring the Echo State Property for stable dynamics~\citep{yildiz2012re}. Only the readout weights $\mathbf{W}_{\text{out}}$ are trained, typically via ridge regression.

The simplicity and training efficiency of ESNs have led to their widespread adoption in time-series prediction~\citep{jaeger2004harnessing}, speech recognition~\citep{verstraeten2007experimental}, and robotic control~\citep{baldini2022reservoir}. Research has focused on optimizing reservoir construction to improve performance and reduce size \citep{DBLP:journals/neco/RodanT12,DINVERNO2025128679,Dale2021}. \cite{DBLP:journals/tnn/RodanT11} proposed the minimum-complexity ESN (MC-ESN) using simple cycles and delays, significantly reducing connection count while maintaining performance. The concept of cycle reservoirs was further explored by~\cite{li2024simple}, and deep reservoir architectures by~\cite{gallicchio2017deep} demonstrated the potential of stacking multiple reservoirs to capture hierarchical temporal features.

Beyond classical ESNs, the RC paradigm encompasses Liquid State Machines~\citep{maass2002real} and Random Neural Networks~\citep{gelenbe1989random}, which share the core RC principle of a fixed untrained recurrent core with a linear readout. Our work, focused on the classic ESN, can be viewed as a step toward bringing sparse efficient learning to the broader RC family.

\subsection{Sparsity in Neural Networks}

Sparsity has long been recognized as a desirable property in machine learning for improving generalization, interpretability, and efficiency~\citep{hastie2015statistical}. L2 regularization penalizes the squared magnitude of weights, producing small but in general non-zero values and thereby smoothing the model. In contrast, L1 regularization penalizes the absolute value of weights, driving many to exactly zero~\citep{tibshirani1996regression}. This makes L1 particularly attractive for feature selection and model compression.

In neural networks, sparsity has been explored through a variety of approaches, including L1-based regularization, pruning techniques, and structured sparsity schemes such as Group LASSO~\citep{tibshirani1996regression,NIPS2015_ae0eb3ee,yuan2006model}. A recent overview is provided by~\citet{10328608}. While effective in centralized settings, many of these approaches either require multiple training phases or are not naturally suited to distributed optimization frameworks.

Within the ESN literature, sparse reservoirs have been shown to reduce computational load without significantly compromising performance~\citep{DBLP:journals/tnn/RodanT11}, and sparse input connections have been explored to reduce input dimensionality~\citep{DBLP:conf/inista/Gallicchio20}. The most direct route to a compact model, however, is to induce sparsity in the readout layer.

\cite{qiao2019adaptive} proposed an Adaptive LASSO ESN that uses the non-differentiable LARS solver to obtain sparse readout weights. While this successfully achieves sparsity, the reliance on a non-gradient-based solver is a critical limitation when extending ESNs to Federated Learning (FL), where the standard algorithm FedAvg~\citep{mcmahan2017} requires clients to compute gradients of a smooth loss function. The two decades since the LASSO was introduced have produced a large body of algorithms for exactly this difficulty, and a comprehensive account of the numerical options is given by the survey of~\cite{zhao2023survey}. Broadly, they fall into three families. Proximal and splitting methods keep the penalty outside the smooth part and handle it through its proximal operator, from iterative shrinkage-thresholding and its accelerated variant~\citep{daubechies2004iterative,beck2009fast} to the general proximal framework of~\cite{parikh2014proximal} and to ADMM~\citep{boyd2011admm}, which is the basis of many distributed LASSO solvers. Subgradient- and stochastic-optimizer approaches instead adapt the training loop itself, as in truncated-gradient online learning~\citep{langford2009sparse} or composite-objective adaptive methods such as AdaGrad with an L1 term~\citep{duchi2011adagrad}. Finally, smoothing methods replace the penalty by a differentiable surrogate, which makes the whole objective amenable to ordinary gradient-based and second-order machinery. \cite{schmidt2007fast} introduced such a twice-differentiable approximation of the L1 norm.

We adopt the third route. This is not a claim that it dominates the others in accuracy or convergence rate; ADMM and accelerated proximal methods are at least as strong for the centralized LASSO, and ADMM in particular has well-developed distributed forms. The reason is the specific requirement of our federated update. A smooth objective is the only one of the three families for which each client can form both a gradient and a Gauss-Newton Hessian of the \emph{complete} objective, penalty included, in a shared coordinate system, so that the server can sum them and take a damped Newton step (Section~\ref{sec4}). Proximal and thresholding schemes place the penalty in an operator that has no additive curvature, and subgradient schemes give first-order information only. The Schmidt surrogate is moreover simple to implement, numerically robust, and has closed-form first and second derivatives, which is what makes the second-order aggregation practical.

\subsection{Sparsity in Signal Analysis and Inverse Problems}

A stronger line of results, compressed sensing, establishes exact recovery of sparse coefficients from few measurements under restricted isometry or incoherence conditions~\citep{candes2006robust,donoho2006compressed}, together with a family of L0-oriented greedy and hard-thresholding solvers (matching pursuit~\citep{mallat1993matching}, orthogonal matching pursuit~\citep{tropp2007signal}, iterative hard thresholding~\citep{blumensath2009iterative}), the exact L0 problem itself being NP-hard~\citep{natarajan1995sparse}. We do not adopt these methods, for three reasons. First, their guarantees rest on incoherence of the design matrix, which reservoir states violate by construction, because all neurons are driven by the same input through contractive dynamics, so the columns of $\mathbf{X}$ are strongly cross-correlated and the L0--L1 equivalence results do not transfer. The same coherence also makes the selected L1 support unstable, in the sense that small perturbations of the data can change which neurons are retained.

Second, that instability is harmless in our setting, because we are not trying to identify a true support. Compressed sensing recovers a signal known to be sparse from scarce measurements; here measurements are plentiful, no ground-truth sparse readout exists, and the goal is model compression, so any sparse and predictive readout is acceptable regardless of which of the correlated neurons carry the weights. Third, and decisively for the federated setting, L0 and greedy methods return discrete support sets that differ across clients and expose neither gradients nor Hessians that add across clients, ruling out the second-order aggregation used here. The smooth penalty instead keeps the optimization in a continuous space where local gradients and Hessians are summable, enabling a Newton-type server update, and defers the discrete decision to a single shrinkage step after aggregation. The smoothing parameter governs a trade-off, in that smaller values approximate the L1 penalty more closely but inflate the curvature near zero, worsening the conditioning of the Hessian.

\subsection{Federated Reservoir Computing}

Federated Learning (FL) is a distributed machine learning paradigm that
enables collaborative model training across decentralized data sources
while preserving data privacy~\citep{mcmahan2017}. Each client performs
local training using its private data and shares only model updates with a
coordinating server, which aggregates them into an improved global model.
This iterative federated optimization cycle keeps raw data on the client and
leverages heterogeneous distributed datasets, making FL
attractive for healthcare, autonomous driving, smart manufacturing, and
collaborative
robotics~\citep{GUO2021108122,PrivacyHealth,10.1145/3533708,2024kishawy,%
ae379c1c7ee843cd9b54cf200b546a11,electronics14071323}.

FL compliance with regulatory frameworks such as GDPR~\citep{EuropeanParliament2016a} comes at the cost of new challenges. Client datasets are often heterogeneous and non-identically distributed, potentially causing client drift~\citep{shi2022optimization}. Communication efficiency is a critical design constraint, addressed through specialized aggregation mechanisms, privacy-preserving techniques (e.g., differential privacy, homomorphic encryption, secure aggregation~\citep{RODRIGUEZBARROSO2020270,fi13040094}), and numerous FL extensions~\citep{LIU2024128019,wenjie22,10.1145/3594300.3594312}. For a broader overview of federated learning, we refer the interested reader to~\citet{DBLP:journals/ijon/PolatoHRS26}.

Recent work has explored distributed ESN implementations. In a foundational contribution, \citet{bacciu2021federated} introduced Federated Reservoir Computing, where clients share only sufficient statistics with the server, achieving a global readout mathematically equivalent to the centralized model. \cite{de2022federated} extended this with Federated Intrinsic Plasticity (FedIP), enabling unsupervised reservoir adaptation via minimal gain and bias parameter exchange. \cite{nogami2025federated} proposed an incremental federated variant for time-series anomaly detection using Mahalanobis Distance of Reservoir States~\citep{tamura2023mahalanobis}. These approaches retain classical regularization methods but lack explicit mechanisms for communication efficiency through sparsity. Our work addresses this gap by applying smooth sparsification to the ESN readout layer.

\section{Methodology}\label{sec3}

\subsection{Echo State Network Fundamentals}

Echo State Networks (ESNs) are a subclass of RNNs that efficiently process temporal data by separating dynamic memory (reservoir) from the trainable output layer~\citep{jaeger2001echo,DBLP:journals/tnn/RodanT11}. An ESN comprises a fixed randomly generated reservoir and a trainable readout.

The input weight matrix $\mathbf{W}_{\text{in}} \in \mathbb{R}^{N_R \times
N_I}$ projects input $\mathbf{u}(t) \in \mathbb{R}^{N_I}$ into the
high-dimensional reservoir space, and the recurrent weight matrix
$\mathbf{W}_{\text{res}} \in \mathbb{R}^{N_R \times N_R}$ evolves the
reservoir state. Leaky integrator units with leaking rate $\gamma \in
(0,1]$ provide temporal smoothing. The state update equations are:
\begin{equation}
\tilde{\mathbf{x}}(t) = \tanh\!\left(\mathbf{W}_{\text{in}} \mathbf{u}(t)
+ \mathbf{W}_{\text{res}} \mathbf{x}(t-1) \right)
\end{equation}
\begin{equation}
\mathbf{x}(t) = (1 - \gamma) \mathbf{x}(t-1) + \gamma \tilde{\mathbf{x}}(t)
\label{eq:leaky}
\end{equation}

These two equations already contain all quantities that have to be chosen. The recurrent matrix $\mathbf{W}_{\text{res}}$ is not trained but drawn at random and then rescaled, and it is this rescaling that introduces the spectral radius. Let $\rho(\mathbf{W}_{\text{res}}) = \max_i |\lambda_i(\mathbf{W}_{\text{res}})|$ be the largest absolute eigenvalue of the sampled matrix. A random draw $\mathbf{W}_0$ is normalized to a target value $\rho$ by setting $\mathbf{W}_{\text{res}} = \rho\,\mathbf{W}_0 / \rho(\mathbf{W}_0)$, so $\rho$ appears in the state update only through the scale of $\mathbf{W}_{\text{res}}\mathbf{x}(t-1)$. It controls how strongly the previous state is fed back and therefore how long the influence of an input persists in the reservoir. Small values of $\rho$ yield fast-fading, nearly memoryless dynamics, whereas values approaching or exceeding one lengthen the memory but risk losing the echo state property, i.e., the requirement that the state asymptotically depends on the input history alone and not on the initial state. The condition $\rho(\mathbf{W}_{\text{res}}) < 1$ is the standard practical rule of thumb for obtaining this property, though it is in general neither necessary nor sufficient; we refer to~\cite{yildiz2012re} for the precise conditions and to~\cite{lukovsevivcius2012practical} for the accompanying tuning guidelines. Similarly, the input scaling $a_{\text{in}}$ sets the amplitude of $\mathbf{W}_{\text{in}}\mathbf{u}(t)$ and thereby how far the $\tanh$ units are driven into their nonlinear regime, the leaking rate $\gamma$ interpolates in Eq.~\eqref{eq:leaky} between retaining the previous state ($\gamma \to 0$) and replacing it entirely ($\gamma = 1$), matching the reservoir timescale to that of the signal, and the reservoir dimension $N_R$ fixes the dimensionality of the feature space seen by the readout.

The reservoir hyperparameters tuned in this work are accordingly the spectral radius $\rho$, the leaking rate $\gamma$, and the input scaling factor $a_{\text{in}}$ (together with the regularization strength $\lambda$ of the readout introduced below); the reservoir dimension is fixed at $N_R = 500$ (Section~\ref{sec5}), and the reservoir connectivity $\phi$ (the fraction of non-zero entries of $\mathbf{W}_{\text{res}}$) is held fixed at $\phi = 0.1$ throughout. The readout layer maps reservoir states to predictions via:
\begin{equation}
\hat{\mathbf{y}}(t) = \mathbf{W}_{\text{out}} \mathbf{x}(t)
\end{equation}
where $\mathbf{W}_{\text{out}} \in \mathbb{R}^{N_O \times N_R}$ is the only trained component and $\mathbf{y}(t)$ denotes the corresponding target. Following standard practices~\citep{DBLP:journals/tnn/RodanT11}, $\mathbf{W}_{\text{in}}$ is sampled uniformly from a range controlled by $a_{\text{in}}$ and $\mathbf{W}_{\text{res}}$ is initialized as a sparse random matrix with spectral radius scaled to $\rho$ to promote the echo state property. In our implementation, we use the ReservoirPy library~\citep{trouvain2020reservoirpy}.

\subsection{Smooth-L1 Regularization for ESN Readout}

While reservoir weights remain fixed, $\mathbf{W}_{\text{out}}$ is trained to minimize a prediction error objective. Sparse readout weights reduce model complexity, enhance generalization, and enable efficient inference. L1 regularization~\citep{tibshirani1996regression} is theoretically ideal for sparsification but introduces a non-differentiable point at zero, precluding second-order optimization. L2 regularization is smooth but does not produce exact sparsity.

We adopt Smooth-L1 regularization~\citep{schmidt2007fast}, which
approximates the L1 norm with a smooth twice-differentiable function:
\begin{equation}
|w|_\alpha = \frac{1}{\alpha} \left[\ln(1 + e^{-\alpha w}) +
\ln(1 + e^{\alpha w})\right]
\end{equation}
This function converges to $|w|$ as $\alpha \to \infty$, remaining smooth
and convex for finite $\alpha \ge 1$. The regularized objectives are:
\begin{equation}
\text{Regression: }\;
\mathcal{L}(\mathbf{W}_{\text{out}}) = \frac{1}{2} \sum_{t}
\left\|\mathbf{y}(t) - \hat{\mathbf{y}}(t)\right\|_2^2 + \lambda
\sum_{i,j} |w_{ij}|_\alpha
\end{equation}
\begin{equation}
\text{Classification: }\;
\mathcal{L}(\mathbf{W}_{\text{out}}) = - \sum_{t} \sum_{c}
y_c(t) \log \hat{y}_c(t) + \lambda \sum_{i,j} |w_{ij}|_\alpha
\end{equation}

where $\lambda > 0$ controls the regularization strength, $y_c(t)$ is the one-hot target for class $c$, and the predicted class probability $\hat{y}_c(t)$ is obtained by applying the softmax function to the readout outputs $\mathbf{W}_{\text{out}}\mathbf{x}(t)$. Because $\mathcal{L}$ is twice differentiable, both objectives can be minimized efficiently via Newton's method with backtracking line search~\citep{boyd2004convex}, yielding improved convergence rates and reduced weight-update variance compared to first-order methods.

In practice, a fixed large $\alpha$ can cause overly aggressive quadratic approximations near zero, as the Smooth-L1 approximation is not appropriately modeled by a quadratic for variables near $0$ when $\alpha$ is large. To address this, a continuation strategy~\citep{schmidt2007fast} is employed. We begin with a small $\alpha$ ($\alpha_0=1$) where the quadratic approximation is appropriate, and take Newton steps or gradient updates between successive increases of the parameter $\alpha$ (multiplying $\alpha$ by a factor $\kappa > 1$ after each Newton iteration; $\kappa = 2$ in all experiments), terminating at a sufficiently large $\alpha$ ($\alpha_{\text{max}}=5\times10^6$) that yields a near-exact L1 penalty. This schedule balances optimization stability with sparsity induction. Both objectives above are implemented in our framework and trained by the same solver. The experimental part of this paper, however, considers classification only, and the regression variant is not discussed in detail; for completeness, its evaluation on two chaotic benchmarks is reported in Appendix~\ref{app:regression}.

\section{Proposed Method}\label{sec4}
This section presents the proposed SL1ESN training framework for learning sparse readout weights in Echo State Networks. Our approach replaces the standard closed-form readout estimation with an iterative optimization procedure that incorporates a twice-differentiable sparsity-promoting regularizer. This formulation preserves compatibility with gradient-based optimization and therefore supports both centralized training and its extension to federated learning. We first describe the method in the centralized setting and then show how it can be adapted to the distributed case.

Two components act together to produce the sparse readout. The Smooth-L1 penalty, sharpened over the continuation schedule in $\alpha$, drives redundant readout coordinates toward zero inside the optimization, so that the surviving coordinates are re-optimized at every Newton step to absorb the contribution of those that vanish. A shrinkage step with threshold $\tau$ then turns the near-zero coordinates into exact zeros, since the smooth surrogate is differentiable everywhere and therefore never attains exact zeros by itself. Model selection is kept separate from this mechanism. The reservoir hyperparameters and the regularization strength $\lambda$ are chosen to maximize validation accuracy, with no sparsity term in the selection objective and with the shrinkage step effectively disabled during the search (negligible threshold $\tau = 10^{-5}$). The fixed threshold $\tau = 0.1$ is applied only when the selected configuration is evaluated, so the sparsity we report in Section~\ref{sec5} is produced by a fixed threshold acting on an accuracy-selected configuration rather than by a sparsity-targeted search. The threshold $\tau$ is not tuned either, since all reported classification results of SL1ESN and FedSL1ESN, centralized and federated, use the same fixed value $\tau = 0.1$, so no per-dataset choice enters the reported sparsity.

\subsection{Centralized Setting}

In the centralized setting, all training data are available on a single server, and the objective is to learn the readout matrix $\mathbf{W}_{\text{out}}$ by minimizing the regularized loss introduced in Section~\ref{sec3}. In contrast to traditional ESN training, where $\mathbf{W}_{\text{out}}$ is obtained in closed form by ridge regression, our approach formulates readout learning as an iterative optimization problem. This is particularly advantageous because it accommodates non-quadratic regularizers, supports incremental or online updates, and provides a natural basis for extensions to distributed optimization settings such as federated learning.

The centralized training procedure consists of two main stages. First, the input sequence is forward-propagated through the fixed reservoir in order to collect the corresponding reservoir states $\mathbf{x}(t)$. These states define the features used by the readout layer. Second, the readout weights are optimized iteratively with respect to the regularized objective. In our case, this optimization is carried out with Newton's method, exploiting the smoothness of the proposed penalty. At each iteration, the gradient $\nabla \mathcal{L}$ and Hessian $\mathbf{H} = \nabla^2 \mathcal{L}$ are computed, the damped linear system \mbox{$(\mathbf{H} + \mu\mathbf{I})\Delta\mathbf{w} = -\nabla \mathcal{L}$} is solved to obtain the search direction, and an Armijo backtracking line search~\citep[Ch.~3]{nocedal2006numerical} is used to determine the step size. The updated weights are then sparsified with the soft-threshold operator $[\mathcal{S}_\tau(\mathbf{W})]_{ij} = \mathrm{sign}(w_{ij})\max(|w_{ij}|-\tau,\,0)$, and the smoothing parameter is advanced along the continuation schedule, $\alpha \leftarrow \min(\alpha\kappa,\,\alpha_{\max})$. The factor $\kappa$, threshold $\tau$, and regularization strength $\lambda$ are treated as predefined hyperparameters; further implementation details are provided in Appendix~\ref{app1}.

Algorithmically, this procedure is related to, but distinct from, the classical Iterative Shrinkage-Thresholding Algorithm (ISTA) and its accelerated variant FISTA~\citep{daubechies2004iterative,beck2009fast}. It retains the shrinkage step $\mathcal{S}_\tau$, but replaces the first-order gradient step by a damped Newton step, available only because the penalty is smooth, and adds the continuation in $\alpha$. Unlike ISTA, where the L1 penalty is handled solely by the proximal operator, here it also enters the objective as the smooth surrogate of strength $\lambda$, so the readout is already being pulled toward a sparse solution while it is being fitted, and the shrinkage step only realizes the exact zeros. Section~\ref{sec:sensitivity} examines how the two interact.

\paragraph{Convergence criterion.} Because exact zeros are produced by the
soft-threshold operator rather than by the smooth objective itself, the
iterates converge to a fixed point of the combined Newton/threshold map at
which the gradient of the smooth objective does not vanish. A conventional
gradient-norm stopping rule is therefore never triggered in practice. We
instead monitor the sparsity pattern, and training stops once $\alpha$ has
reached $\alpha_{\max}$ and the non-zero support of
$\mathbf{W}_{\text{out}}$ has changed by less than a tolerance
$\varepsilon_s$ over the last $P$ iterations ($P = 8$,
$\varepsilon_s = 0.2$ percentage points in all experiments). This
support-stagnation rule terminates training well within the fixed iteration
budget at unchanged accuracy.
Algorithm~\ref{alg:centralized} summarizes the resulting centralized training procedure.
\begin{algorithm}[htbp]
\caption{Centralized Training of ESN Readout with Smooth-L1
Regularization}\label{alg:centralized}
\begin{algorithmic}[1]
\State \textbf{Input}: Training data
  $\{(\mathbf{u}(t), \mathbf{y}(t))\}_{t=1}^T$,
  reservoir parameters $\mathbf{W}_{\text{in}}, \mathbf{W}_{\text{res}},
  \gamma$, regularization strength $\lambda$, schedule
  $\alpha_0, \kappa, \alpha_{\max}$, threshold $\tau$,
  early-stop window $P$ and tolerance $\varepsilon_s$
\State \textbf{Output}: Optimized readout weights $\mathbf{W}_{\text{out}}$
\State Initialize $\mathbf{W}_{\text{out}} \leftarrow \mathbf{0}$
\State Compute reservoir states $\mathbf{x}(t)$ for all $t$
\State $\alpha \gets \alpha_0$
\While{iteration budget not exhausted}
  \State $\mathbf{g} \gets \nabla_{\mathbf{W}_{\text{out}}} \mathcal{L}$
    (using current $\alpha$)
  \State $\mathbf{H} \gets \nabla^2_{\mathbf{W}_{\text{out}}} \mathcal{L}$
  \State Solve $(\mathbf{H} + \mu\mathbf{I})\,\Delta\mathbf{w} = -\mathbf{g}$
    \Comment{damped Newton direction}
  \State Find step size $\eta$ via backtracking line search
  \State $\mathbf{W}_{\text{out}} \gets \mathcal{S}_\tau\!\left(
    \mathbf{W}_{\text{out}} + \eta\,\Delta\mathbf{w}\right)$
    \Comment{soft-threshold: $\mathrm{sign}(w)\max(|w|-\tau,0)$}
  \State $\alpha \gets \min(\alpha \cdot \kappa,\,\alpha_{\max})$
    \Comment{continuation schedule}
  \If{$\alpha = \alpha_{\max}$ and non-zero support of
    $\mathbf{W}_{\text{out}}$ changed by $< \varepsilon_s$ over the last
    $P$ iterations}
    \State \textbf{break} \Comment{support-stagnation convergence}
  \EndIf
\EndWhile
\State \Return $\mathbf{W}_{\text{out}}$
\end{algorithmic}
\end{algorithm}

\subsection{Distributed Setting}

The distributed training pipeline is illustrated in Figure~\ref{fig:sparse_readout}. We consider a parameter-server architecture with one central server and \(K\) clients, each holding private time-series data. The goal is to learn a global readout matrix \(\mathbf{W}_{\text{out}}\) without sharing raw data, while exploiting the twice-differentiable structure of the Smooth-L1 penalty to perform second-order (Newton) updates on the server and to obtain a sparse global readout for efficient storage and inference. Because the smooth penalty exposes a well-defined gradient and Hessian, we implement the distributed training as a federated Newton scheme rather than the first-order FedAvg averaging of local weights. We refer to this federated variant as FedSL1ESN, reserving SL1ESN for the centralized model described above.

The procedure consists of four main steps:

\begin{enumerate}
    \item \textbf{Initialization:} The server randomly initializes the fixed reservoir
        parameters \(\mathbf{W}_{\text{in}}\) and \(\mathbf{W}_{\text{res}}\) and
        broadcasts them to all clients. The global readout is initialized as
        \(\mathbf{W}_{\text{out}}^{(0)} = \mathbf{0}\) and the smoothing
        parameter as \(\alpha = \alpha_0\).

    \item \textbf{Local curvature computation:} Each client \(k\) receives the
        current global readout \(\mathbf{W}_{\text{out}}^{(t)}\) and, using the
        frozen shared reservoir, evaluates the local gradient
        \(\mathbf{g}_k = \nabla \mathcal{L}_k\) and the per-class Gauss-Newton
        Hessian blocks \(\mathbf{H}_k = \nabla^2 \mathcal{L}_k\) of its
        Smooth-L1 regularized loss at \(\mathbf{W}_{\text{out}}^{(t)}\). No
        local weight update is performed; the client returns
        \((\mathbf{g}_k, \mathbf{H}_k)\). Reservoir states are
        \(\mathbf{W}_{\text{out}}\)-independent and therefore cached after the
        first round.

    \item \textbf{Newton-step aggregation:} The server sums the client
        gradients and Hessians,
        \(\mathbf{g} = \sum_{k}\mathbf{g}_k\) and
        \(\mathbf{H} = \sum_{k}\mathbf{H}_k\), solves a damped Newton system
        per output class \(c\) to form the descent direction
        \(\mathbf{d}_c = -(\mathbf{H}_c + \mu\mathbf{I})^{-1}\mathbf{g}_c\),
        and stacks these rows into the full direction matrix
        \(\mathbf{D} = [\mathbf{d}_1, \dots, \mathbf{d}_{N_O}]^\top\). It then
        selects a step size \(\eta\) by Armijo backtracking line search~\citep{nocedal2006numerical} on the aggregated loss \(\sum_{k}\mathcal{L}_k\). The readout is
        updated and sparsified with a soft-threshold operator
        \(\mathcal{S}_\tau(\cdot)\):
        \[
        \mathbf{W}_{\text{out}}^{(t+1)} =
        \mathcal{S}_\tau\!\Bigl(
        \mathbf{W}_{\text{out}}^{(t)} + \eta\,\mathbf{D}\Bigr),
        \qquad
        [\mathcal{S}_\tau(\mathbf{W})]_{ij} =
        \mathrm{sign}(w_{ij})\max(|w_{ij}|-\tau,\,0).
        \]

    \item \textbf{Broadcasting:} The updated sparse global readout
        \(\mathbf{W}_{\text{out}}^{(t+1)}\) is broadcast to all clients in CSR
        form, the smoothing parameter is advanced
        (\(\alpha \leftarrow \min(\alpha\kappa, \alpha_{\max})\)), and the
        process repeats for the next communication round.
\end{enumerate}

The continuation in \(\alpha\) makes the penalty a progressively sharper approximation of L1 over rounds, so the soft-threshold drives an increasing fraction of readout weights to exactly zero; combined with the Newton step, the global readout converges within a few rounds to a sparse solution. The communication implications of transmitting curvature, together with the storage and inference benefits of the resulting sparse readout, are analyzed in Section~\ref{sec:comm}. The complete distributed training procedure is summarized in Algorithm~\ref{alg:dist_smoothl1_esn}.

\begin{algorithm}[t]
\caption{FedSL1ESN: Distributed (Federated Newton) Training}\label{alg:dist_smoothl1_esn}
\begin{algorithmic}[1]
\State \textbf{Input}: Number of clients $K$, rounds $R$, regularization
  strength $\lambda$, initial smoothing $\alpha_0$, factor $\kappa$,
  threshold $\tau$
\State \textbf{Server executes:}
\State \quad Initialize $\mathbf{W}_{\text{in}}$, $\mathbf{W}_{\text{res}}$
  and broadcast to all clients
\State \quad Initialize $\mathbf{W}_{\text{out}}^{(0)} \leftarrow \mathbf{0}$;
  set $\alpha \gets \alpha_0$
\For{each round $t = 0, 1, \dotsc, R-1$}
  \For{each client $k$ \textbf{in parallel}}
    \State $(\mathbf{g}_k, \mathbf{H}_k) \leftarrow
      \text{ClientUpdate}(k,\,\mathbf{W}_{\text{out}}^{(t)},\,\alpha)$
  \EndFor
  \State $\mathbf{g} \gets \sum_{k} \mathbf{g}_k$,\quad
    $\mathbf{H} \gets \sum_{k} \mathbf{H}_k$
    \Comment{aggregate gradient and per-class Hessian}
  \State Solve damped Newton direction $\mathbf{D}$:
    $\mathbf{d}_c = -(\mathbf{H}_c + \mu\mathbf{I})^{-1}\mathbf{g}_c$ for each
    class $c$
  \State Choose step $\eta$ by Armijo backtracking line search on
    $\sum_{k}\mathcal{L}_k$
  \State $\mathbf{W} \gets \mathbf{W}_{\text{out}}^{(t)} + \eta\,\mathbf{D}$
  \State $\mathbf{W}_{\text{out}}^{(t+1)} \gets \mathcal{S}_\tau(\mathbf{W})$
    \Comment{soft-threshold: $\mathrm{sign}(w)\max(|w|-\tau,0)$}
  \State Broadcast $\mathbf{W}_{\text{out}}^{(t+1)}$ (sparse, CSR) to all
    clients
  \State $\alpha \gets \min(\alpha\cdot\kappa,\,\alpha_{\max})$
    \Comment{continuation schedule}
\EndFor
\Statex
\Statex \textbf{ClientUpdate}$(k,\,\mathbf{W},\,\alpha)$:
\State Compute frozen reservoir states $\mathbf{X}_k$ (cached after round 0)
\State $\mathbf{g}_k \gets \nabla_{\mathbf{W}}\mathcal{L}_k(\mathbf{W})$;\quad
  $\mathbf{H}_k \gets$ per-class Gauss-Newton Hessian of $\mathcal{L}_k$ at
  $\mathbf{W}$
\State \Return $(\mathbf{g}_k, \mathbf{H}_k)$ to server
\end{algorithmic}
\end{algorithm}

\begin{table}[ht]
\caption{Classification datasets. For ease of reference, the \textit{DistalPhalanxOutlineCorrect} dataset is abbreviated as Distal., and Character denotes the \textit{Character Trajectories} dataset from the UCI repository. All datasets are used with the train/test partition of their archive, except Character, for which the UCI repository provides none: its $2858$ sequences are split $80/20$ once, and the same split is reused by every method and seed. Length is the padded sequence length, and variable-length sequences (Character, JapaneseVowels) are zero-padded to the longest sequence in the dataset.}
\label{tab:dataset_classification}
\centering
\begin{tabular}{lrrrrr}
\toprule
\textbf{Dataset} & \textbf{Train} & \textbf{Test} & \textbf{Length} &
  \textbf{Dim.} & \textbf{Classes} \\
\midrule
UCI HAR        & 7352 & 2947 & 128 &  9 & 6 \\
Character      & 2286 &  572 & 205 &  3 & 20 \\
JapaneseVowels &  270 &  370 &  29 & 12 & 9 \\
ECG5000        &  500 & 4500 & 140 &  1 & 5 \\
Distal.        &  600 &  276 &  80 &  1 & 2 \\
Yoga           &  300 & 3000 & 426 &  1 & 2 \\
Strawberry     &  613 &  370 & 235 &  1 & 2 \\
\bottomrule
\end{tabular}
\end{table}

\section{Experiments}\label{sec5}
This section evaluates the proposed SL1ESN on time-series classification. The experiments are designed to assess predictive performance, the degree of sparsity induced in the readout layer, and the behavior of the method in centralized and distributed training scenarios.
\subsection{Datasets}
We use seven datasets from the UCI and UCR archives (Table~\ref{tab:dataset_classification}), spanning univariate and multivariate sequences in areas such as activity recognition~\citep{Anguita2013HAR}, character trajectories~\citep{Williams2006Character}, speech analysis~\citep{kudo1999multidimensional}, and other fields~\citep{dau2019ucr}. The datasets cover binary and multi-class problems across various training sizes and sequence lengths.

\subsection{Centralized setup}
All ESN configurations use a fixed reservoir of size $N_R = 500$ for comparability across regularization methods. Input matrix entries are drawn from a Uniform distribution $\mathcal{U}(-a_{\text{in}}, a_{\text{in}})$; reservoir adjacency matrix entries from $\mathcal{U}(-1, 1)$. The hyperbolic tangent activation confines reservoir outputs to $[-1, 1]$. Hyperparameters are optimized via random search over the distributions in Table~\ref{tab:hyperparameters}. The search objective is predictive accuracy only, namely the validation accuracy, with no sparsity term. During the search the shrinkage threshold is held at the negligible value $10^{-5}$, so the candidate readouts are effectively dense; the fixed threshold $\tau = 0.1$ is applied only when the selected SL1ESN configuration is evaluated. The readout sparsity we report below is therefore produced by this fixed threshold acting on an accuracy-selected configuration, not a quantity the search was asked to maximize. Full per-dataset selected values are reported in Appendix~\ref{app1}.
We compare three readout penalties, all trained with the same damped Gauss--Newton solver so that only the penalty differs and no solver confound enters the comparison: an unregularized cross-entropy readout (none), an L2-regularized readout (L2-ESN), and the proposed SL1ESN. The two dense baselines are run with the same negligible threshold $10^{-5}$ in both search and evaluation, which leaves them dense. Reported accuracies are means $\pm$ standard deviation over ten independent random seeds. The softmax parameterization is shift invariant, so the unregularized Hessian is singular along the all-ones direction; rather than fixing a reference class we retain the full $N_O \times N_R$ readout and rely on the adaptive damping $\mu\mathbf{I}$ of the Newton solve (Appendix~\ref{app1}) to render the system non-singular, which keeps the parameterization identical across the three penalties. Readout sparsity is accordingly always reported as the fraction of zeros among all $N_O N_R$ entries, none of which is constrained to zero a priori.
\begin{table}[ht]
    \centering
\caption{Hyperparameter search distributions (random search).}
\label{tab:hyperparameters}
\begin{tabular}{ll}
\toprule
\textbf{Parameter} & \textbf{Distribution} \\
\midrule
Spectral radius ($\rho$)    & Uniform$(0.3,\ 3.0)$ \\
Input scaling ($a_{\text{in}}$) & Log-uniform$[10^{-2},\ 10^{2}]$ \\
Leaking rate ($\gamma$)     & Log-uniform$[5\times10^{-3},\ 1]$ \\
Regularization ($\lambda$)  & Log-uniform$[10^{-4},\ 10^{2}]$ \\
\bottomrule
\end{tabular}
\end{table}

\subsection{Distributed setup}\label{sec:dist_config}

All federated experiments use the following fixed configuration: $K = 5$ clients and $50$ communication rounds, with results averaged over ten seeds (a sensitivity study over the client count $K$ is reported in Appendix~\ref{app2}). Each round follows the federated Newton scheme of Algorithm~\ref{alg:dist_smoothl1_esn}: clients return the local gradient and per-class Gauss-Newton Hessian of their Smooth-L1 loss, and the server applies a single damped Newton step with Armijo backtracking line search ($c_1 = 10^{-4}$, shrinkage $0.5$) followed by a soft-threshold $\mathcal{S}_\tau$. The smoothing parameter follows the continuation schedule with $\alpha_0 = 1$ and factor $\kappa = 2$. Data are partitioned uniformly (IID) across clients; heterogeneous (non-IID) partitions are not implemented in this study, and Section~\ref{sec:limitations} discusses this limitation together with the existing line of Newton-type federated methods for heterogeneous data that a future version can build on. The global reservoir parameters ($\mathbf{W}_{\text{in}}$, $\mathbf{W}_{\text{res}}$) are initialized once on the server and broadcast to all clients at the start; they remain fixed throughout all rounds. As in the centralized setting, hyperparameters are selected for accuracy alone. The proposed FedSL1ESN is compared against L2 and unregularized ESN baselines,
all trained with the same server-side Newton update so that the readout penalty
is the only variable that differs, which makes the unregularized column a
federated second-order baseline in its own right. No first-order FedAvg
comparison is included, for a structural reason: clients perform no local weight
update but return $(\mathbf{g}_k, \mathbf{H}_k)$ evaluated at the current global
readout, and under full participation the data-fidelity part of their sums is exactly the gradient and Gauss-Newton Hessian of the pooled cross-entropy loss, independently of how the samples are partitioned. The federated iteration is therefore a centralized Newton iteration on an objective that differs from the centralized one only in the penalty, which every client carries a copy of, so that the aggregate is $K\lambda$ rather than $\lambda$. Two further differences keep Table~\ref{tab:fl_results} from reproducing Table~\ref{tab:acc_spar_classification} numerically: the federated runs use a fixed budget of $50$ rounds where the centralized solver iterates to the support-stagnation criterion of Section~\ref{sec4}, and $\lambda$ is re-selected for the federated setting (Appendix~\ref{app1}). Against that background FedAvg, which optimizes the same convex objective by first-order steps, could only approach the same optimum more slowly and without producing a sparse readout, which is why it is not included as a separate column. The server broadcasts the global readout in compressed sparse row (CSR) form; the resulting sparsity benefits for model storage and inference are discussed in Section~\ref{sec:comm}. Although the readout typically reaches near-peak accuracy within a few rounds (Figure~\ref{fig:acc_vs_rounds}), we report at $50$ rounds so that the continuation schedule has fully sharpened the penalty and the readout sparsity has stabilized.

\subsection{Results and Analysis}

\subsubsection{Centralized setting}

Table~\ref{tab:acc_spar_classification} reports the test accuracy of the three readout penalties together with the readout sparsity that SL1ESN attains at the fixed threshold $\tau = 0.1$. Classification is where the sparse readout pays off: SL1ESN discards $77$--$95\%$ of the readout weights while remaining close to the dense baselines in accuracy. It is the most accurate method on UCI HAR, and on Character, JapaneseVowels, ECG5000, and Distal.\ its ten-seed mean accuracy is within $0.7$ points of that of the better of the unregularized and L2 readouts. On Distal.\ and JapaneseVowels this difference is smaller than the seed-to-seed standard deviation, whereas on Character and ECG5000 it is larger, so there the sparse readout is slightly less accurate. These are differences of means over our seeds that do not account for seed-to-seed variation, and they should not be read as bounds that hold for other seeds.

\begin{figure*}[htbp]
\centering
\includegraphics[width=0.8\textwidth]{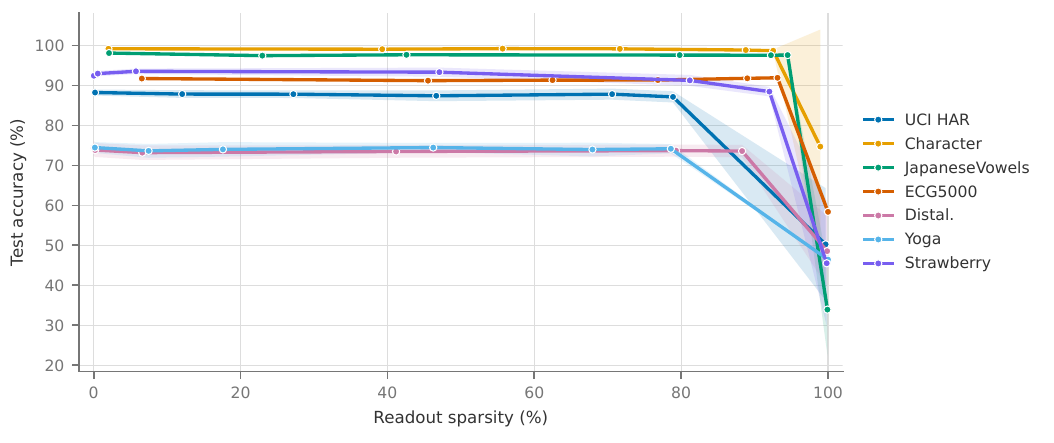}
\caption{\textbf{Centralized setting.} Test accuracy versus readout sparsity for the accuracy-selected SL1ESN configuration, trained on the pooled data by a single learner (mean over ten seeds), obtained by sweeping the shrinkage threshold $\tau$; shaded bands are $\pm$1 std across seeds. Accuracy stays essentially flat up to $80$--$95\%$ sparsity on every dataset, and collapses only once the readout is driven to near-total sparsity. Figure~\ref{fig:fl_acc_vs_sparsity} repeats this sweep for the federated global readout; the two are plotted on the same axes so that the curves can be compared directly, and their similarity is itself the result, indicating that federation does not degrade the sparsity--accuracy trade-off.}
\label{fig:acc_vs_sparsity_thres}
\end{figure*}

Two datasets, Yoga and Strawberry, trail the unregularized readout by $5.8$ and $5.7$ points. Figure~\ref{fig:acc_vs_sparsity_thres} shows that this is not a cost of sparsification. It plots test accuracy against readout sparsity as the shrinkage threshold is swept over the accuracy-selected configuration, and on all seven datasets, Yoga and Strawberry included, the curve is essentially flat out to $80$--$95\%$ sparsity before it collapses at near-total sparsity. On Yoga and Strawberry the gap to the unregularized readout is already present at the left edge of the plot, where the readout is barely sparse at all (at $0.2\%$ sparsity Yoga is already $5.1$ points below its unregularized baseline), and it stays essentially constant as sparsity grows. In other words, the Smooth-L1-trained readout is itself a little weaker than the unregularized one on these two datasets, independently of how much of it is subsequently pruned. That the plateau is present everywhere is what makes a single global threshold viable: most of the reservoir readout is genuinely redundant for classification, and removing it costs nothing measurable until almost none of it is left.

\begin{table*}[ht]
\centering
\caption{Centralized classification test accuracy (\%) and SL1ESN
readout sparsity. All readouts are selected for predictive accuracy;
\textbf{bold} marks the highest accuracy per dataset. Sparsity is the fraction
of zero readout weights; the unregularized and L2 readouts have no
sparsification mechanism and remain dense. All values are mean $\pm$ std over
ten seeds. A single fixed shrinkage threshold $\tau = 0.1$ is used for every
dataset.}
\resizebox{0.7\textwidth}{!}{%
\begin{tabular}{lccccc}
\toprule
\multirow{2}{*}{\textbf{Dataset}} &
  \textbf{none} & \textbf{L2-ESN} &
  \multicolumn{2}{c}{\textbf{SL1ESN (ours)}} \\
  \cmidrule(lr){2-2}\cmidrule(lr){3-3}\cmidrule(lr){4-5}
  & Acc.\ (\%) & Acc.\ (\%) & Acc.\ (\%) & Sparsity (\%) \\
\midrule
UCI HAR        & $87.37\pm1.11$ & $88.01\pm0.43$ &
  $\mathbf{88.05}\pm0.90$ & $77.6\pm1.5$ \\
Character      & $\mathbf{99.58}\pm0.12$ & $98.88\pm0.28$ &
  $98.97\pm0.25$ & $92.7\pm0.6$ \\
JapaneseVowels & $\mathbf{98.24}\pm0.63$ & $97.57\pm0.17$ &
  $97.68\pm0.56$ & $94.6\pm0.4$ \\
ECG5000        & $91.38\pm0.64$ & $\mathbf{92.37}\pm0.08$ &
  $91.91\pm0.20$ & $93.3\pm0.4$ \\
Distal.        & $\mathbf{73.55}\pm0.81$ & $73.08\pm1.23$ &
  $73.51\pm1.52$ & $89.6\pm3.3$ \\
Yoga           & $\mathbf{79.53}\pm0.94$ & $79.24\pm2.38$ &
  $73.70\pm1.60$ & $77.4\pm2.4$ \\
Strawberry     & $\mathbf{96.11}\pm0.64$ & $93.68\pm0.96$ &
  $90.38\pm1.27$ & $88.6\pm2.0$ \\
\bottomrule
\end{tabular}%
}
\label{tab:acc_spar_classification}
\end{table*}

\begin{table*}[ht]
\centering
\caption{Federated learning results (five clients, $50$ rounds, IID split,
mean $\pm$ std over ten seeds; test accuracy) and FedSL1ESN readout
sparsity, all reported at the final round. All readouts are selected for
predictive accuracy; \textbf{bold} marks the highest accuracy per dataset. The
unregularized and L2 readouts have no sparsification mechanism and remain dense
($<2\%$ zeros). The same fixed shrinkage threshold $\tau = 0.1$ as in the
centralized setting is used for every dataset.}
\label{tab:fl_results}
\resizebox{0.7\textwidth}{!}{%
\begin{tabular}{lcccc}
\toprule
\multirow{2}{*}{\textbf{Dataset}} &
  \textbf{none} & \textbf{L2-ESN} &
  \multicolumn{2}{c}{\textbf{FedSL1ESN (ours)}} \\
  \cmidrule(lr){2-2}\cmidrule(lr){3-3}\cmidrule(lr){4-5}
  & Acc.\ (\%) & Acc.\ (\%) & Acc.\ (\%) & Sparsity (\%) \\
\midrule
UCI HAR & $\mathbf{89.41}\pm1.15$ & $87.79\pm0.72$ & $87.86\pm0.79$ & $74.9\pm0.5$ \\
Character & $98.48\pm0.35$ & $\mathbf{98.74}\pm0.26$ & $98.20\pm0.57$ & $84.8\pm2.2$ \\
JapaneseVowels & $\mathbf{96.19}\pm0.96$ & $81.27\pm2.12$ & $94.84\pm1.33$ & $95.2\pm0.3$ \\
ECG5000 & $90.32\pm0.32$ & $\mathbf{91.38}\pm0.77$ & $91.30\pm0.34$ & $91.6\pm1.0$ \\
Distal. & $67.10\pm2.13$ & $71.93\pm1.33$ & $\mathbf{73.77}\pm2.56$ & $86.7\pm2.4$ \\
Yoga & $71.60\pm1.32$ & $\mathbf{73.45}\pm1.42$ & $69.00\pm1.82$ & $80.8\pm1.9$ \\
Strawberry & $83.41\pm2.54$ & $\mathbf{91.32}\pm1.19$ & $88.89\pm1.81$ & $89.2\pm3.0$ \\
\bottomrule
\end{tabular}%
}
\end{table*}

\begin{figure*}[htbp]
\centering
\includegraphics[width=0.8\textwidth]{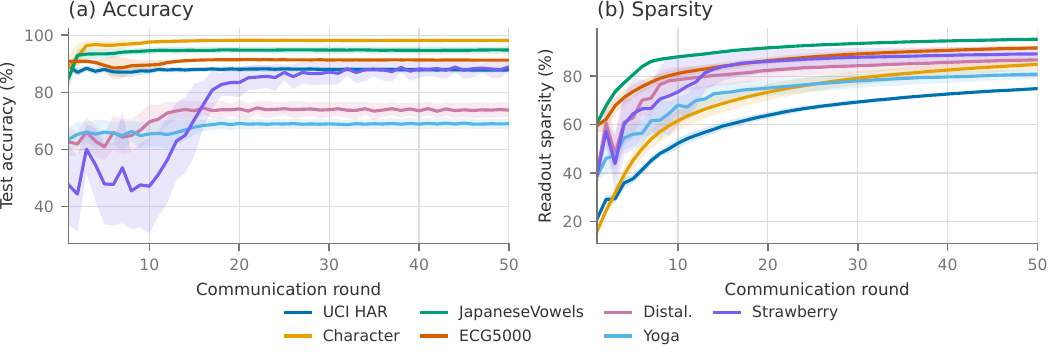}
\caption{FedSL1ESN convergence over communication rounds for each dataset
(five clients, IID split, $50$ rounds, mean over ten seeds; shaded bands are
$\pm$1 std). (a)~Test accuracy reaches near-peak within a few rounds on most
datasets and is stable thereafter (Strawberry is the slowest, climbing until
round $\sim20$). (b)~Readout sparsity accrues more gradually over the same
rounds, as the continuation schedule sharpens the penalty toward L1, and has
largely levelled off by round $50$; this is why we fix the budget at $50$
rounds and report there. The two panels separate the two timescales of the
method, since the cross-entropy fit is reached almost immediately, while the
sparsification unfolds over the full round budget. The bands in (b) are wide at
early rounds, where seeds sit at different points on the steep part of the curve, and narrow by round $50$, which is why the round-$50$ sparsity is
reproducible across seeds.}
\label{fig:acc_vs_rounds}
\end{figure*}

\begin{figure*}[htbp]
\centering
\includegraphics[width=0.8\textwidth]{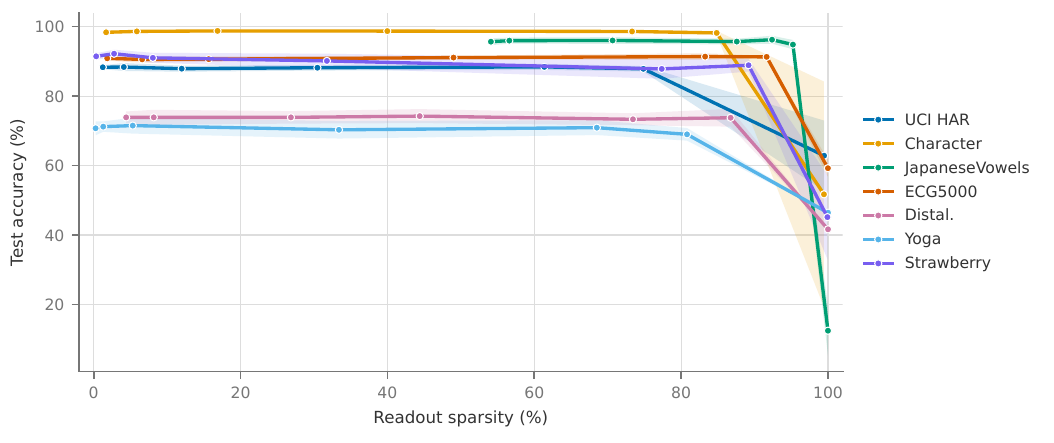}
\caption{\textbf{Federated setting; the counterpart of Figure~\ref{fig:acc_vs_sparsity_thres}.} Test accuracy versus readout sparsity for the accuracy-selected \emph{global} readout obtained by FedSL1ESN, i.e., the model aggregated from five clients that never share data ($50$ rounds, IID partition, mean over ten seeds, reported at the final round), under the same sweep of the shrinkage threshold $\tau$; shaded bands are $\pm$1 std across seeds. As in the centralized case, accuracy stays essentially flat out to $75$--$95\%$ sparsity on most datasets, and collapses only once the readout is driven to near-total sparsity.}
\label{fig:fl_acc_vs_sparsity}
\end{figure*}

\subsubsection{Distributed setting}

Table~\ref{tab:fl_results} reports the federated results at the final round (round $50$), obtained with the same fixed threshold $\tau = 0.1$. The picture from the centralized experiments carries over, that FedSL1ESN learns a global readout that is $75$--$95\%$ sparse while tracking the dense baselines closely in accuracy. It is the most accurate method on Distal., within $0.6$ points of the best on Character and ECG5000, and within $1.6$ points on UCI HAR; on JapaneseVowels it is within $1.4$ points of the unregularized readout and outperforms the L2 readout by a wide margin ($94.84\%$ against $81.27\%$), the federated L2 solution being the one configuration in our study that fails outright. On Character and ECG5000 the margin to the best readout is smaller than the seed-to-seed standard deviation, and on JapaneseVowels it is comparable to it, so the sparse global readout stays close to the best dense one on five of the seven benchmarks.
That failure is not an unlucky choice of $\lambda$: across the whole searched range, from $10^{-4}$ to $10^{2}$, no federated L2 configuration on JapaneseVowels exceeds $80\%$ selection accuracy, whereas the unregularized and Smooth-L1 readouts reach $100\%$ under the identical protocol, so the penalty rather than its strength is what breaks. JapaneseVowels is also the smallest benchmark, so at $K = 5$ each client holds roughly $54$ sequences and returns a Gauss-Newton block of rank at most $54$ against $N_R = 500$; unlike the Smooth-L1 curvature, which decays for $|w| \gg 1/\alpha$ and therefore leaves the large, resolved coordinates almost free, the isotropic $\lambda\mathbf{I}$ that L2 contributes acts equally on every direction, resolved or not, and is replicated $K$ times by the aggregation. We report this as an observation rather than a fully diagnosed mechanism. As centrally, Yoga and Strawberry are the datasets that pay for the uniform threshold, trailing the best baseline by $4.5$ and $2.4$ points. The unregularized and L2 readouts offer no comparable saving in return, since they possess no sparsification mechanism at all and stay dense ($<2\%$ zeros even under L2 shrinkage).

Reporting at the final round matters here, and Figure~\ref{fig:acc_vs_rounds} shows why. Two timescales are at work: test accuracy reaches near-peak within a few communication rounds on most datasets (panel a), a direct consequence of the second-order server update, whereas readout sparsity keeps accruing over more rounds as the continuation schedule sharpens the penalty toward L1 (panel b). The sparsity therefore has not settled at the round where validation accuracy first peaks; only by round $50$ has it largely levelled off, which is why we fix the budget at $50$ rounds and report there. This also makes the sparsity reproducible across seeds: the standard deviation of the sparsity column in Table~\ref{tab:fl_results} is at most $3$ points on every dataset, because at round $50$ the readouts have converged onto the flat part of their sparsification curves, rather than being sampled while still climbing. As a further check, the accuracy--sparsity curve of the global readout (Figure~\ref{fig:fl_acc_vs_sparsity}) has the same flat plateau as its centralized counterpart, so the sparsity is not an artifact of aggregating over few clients but a property of the readout the federated optimization converges to.

\subsubsection{Ablation and Sensitivity Analysis}\label{sec:sensitivity}

\begin{figure*}[htbp]
  \centering
  \includegraphics[width=0.8\textwidth]{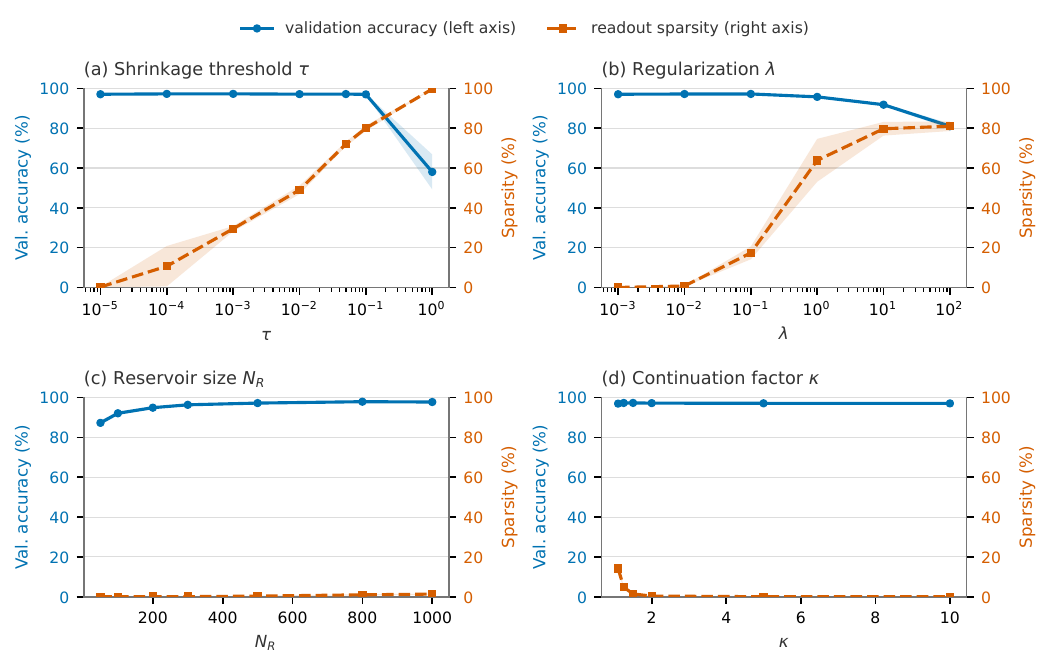}
  \caption{One-at-a-time ablation on UCI HAR (mean over ten seeds; shaded bands
    are $\pm$1 std). Left axis
    (blue): validation accuracy; right axis (orange): readout sparsity.
    (a)~The shrinkage threshold $\tau$ trades sparsity for accuracy cleanly.
    (b)~The penalty strength $\lambda$ also sparsifies, but sacrifices more
    accuracy and saturates. (c)~Reservoir size $N_R$ governs accuracy and
    saturates around $500$. (d)~The continuation factor $\kappa$ leaves
    accuracy unchanged.}
  \label{fig:ablation}
\end{figure*}

To isolate the role of each hyperparameter we perform a one-at-a-time ablation on the UCI HAR dataset, varying a single parameter over a grid while holding the others at their selected values and averaging validation accuracy and readout sparsity over random seeds. Figure~\ref{fig:ablation} reports the four parameters that matter for the accuracy--sparsity trade-off. Note that the accuracies here are measured on the validation split used for model selection and therefore sit well above the held-out test accuracies of Table~\ref{tab:acc_spar_classification}; only the relative trends are of interest.

Sweeping the shrinkage threshold $\tau$ drives sparsity smoothly from $0$ to $100\%$ while validation accuracy stays essentially flat ($\approx 97\%$) up to $\sim80\%$ sparsity, collapsing only once the readout is driven to near-total sparsity. The regularization strength $\lambda$ also induces sparsity in classification, but on markedly worse terms. At matched $\sim65\%$ sparsity, $\lambda=1$ costs about $1.5$ accuracy points where the threshold costs none, and beyond $\lambda\approx10$ accuracy collapses while sparsity plateaus near $83\%$. The two mechanisms are therefore not interchangeable, and this is why the search is free to optimize $\lambda$ for accuracy alone. At the selected values it stays small, and the sparsity in Table~\ref{tab:acc_spar_classification} is produced by the penalty and shrinkage acting together during training rather than by a large $\lambda$.

The remaining two parameters do not affect sparsity. Accuracy rises with the reservoir size $N_R$ and saturates beyond $N_R\approx500$ (from $\approx88\%$ at $N_R=50$ to $\approx98\%$ at $N_R=800$), while readout sparsity stays near zero throughout, confirming that $N_R$ governs representational capacity and motivating the fixed choice $N_R=500$ as a cost-accuracy saturation point. Accuracy is likewise essentially invariant to the continuation factor $\kappa$, with negligible residual sparsity for $\kappa\ge2$ (a small effect appears only for very slow schedules such as $\kappa=1.1$); the initial smoothing $\alpha_0$ is equally inconsequential (not shown). The continuation schedule therefore needs no per-dataset tuning.

\section{Discussion}\label{sec6}

\subsection{Sparsity--Accuracy Trade-off}

The central finding is that on time-series classification the ESN readout is largely redundant, and that SL1ESN is able to find and discard that redundancy at little cost in accuracy. Under a single threshold shared by every dataset and both settings, its readouts are $77$--$95\%$ sparse centrally and $75$--$95\%$ sparse in the federated setting, and on the same five of the seven classification benchmarks their ten-seed mean accuracy is within $0.7$ points of that of the strongest dense baseline centrally and within $1.6$ points in federation. These are differences of means over our seeds; in most cases they are smaller than the seed-to-seed standard deviation, so on these benchmarks the sparse readout is close to the best dense one rather than demonstrably equal to it. On UCI HAR the centralized sparse readout is the most accurate model outright, and in federation that holds on Distal. The accuracy--sparsity curves of Figures~\ref{fig:acc_vs_sparsity_thres} and~\ref{fig:fl_acc_vs_sparsity} show why this is possible, they are flat over a wide range before they break, so there is a broad band of sparsity levels at which the readout simply loses nothing. This is the payoff of the smooth penalty. Because the surrogate is differentiable, the redundant coordinates are driven toward zero while the readout is still being fitted, and the coordinates that survive are re-optimized at every Newton step to absorb what the others gave up. A post-hoc magnitude pruning of a finished dense fit cannot do this.

The concern that highly separable datasets might resist sparsification does not materialize. JapaneseVowels, a $12$-dimensional $9$-class problem with short sequences, sparsifies to $94.6\%$ centrally and $95.2\%$ in federation while remaining within $0.6$ and $1.4$ points of the best baseline. Where SL1ESN does give up accuracy is Yoga and Strawberry, which trail the best baseline by $5.8$ and $5.7$ points centrally and by $4.5$ and $2.4$ points in federation. In the centralized setting Figure~\ref{fig:acc_vs_sparsity_thres} localizes the cause, on these two datasets the sparsity plateau is just as flat as on the others, and the gap is already present at the dense end of the sweep, so it comes from the Smooth-L1-trained fit being slightly weaker than the unregularized one, not from the pruning, and lowering $\tau$ would therefore not recover it. The federated setting behaves differently on exactly these two datasets: the accuracy--sparsity curve does slope down (Yoga loses $1.7$ points and Strawberry $2.6$ between their dense and their $\tau=0.1$ operating points), so there part of the shortfall is a genuine cost of sparsification. This is the one place where the centralized and federated readouts diverge, and it coincides with the two noisiest, smallest-margin datasets in the suite.

\subsection{Communication Efficiency}\label{sec:comm}

The sparsity of the FedSL1ESN global readout yields efficiency gains along three axes, which we state precisely to avoid over-claiming. First, model storage and inference. The converged global readout has only $5$--$25\%$ of its entries non-zero, so the stored model and the per-sample inference cost (a sparse matrix--vector product on each edge device) shrink by the corresponding factor. This is the benefit most relevant to the resource-constrained IoT setting that motivates this work. Second, the server-to-client broadcast, which carries the global readout and is transmitted in CSR form and therefore inherits the same compression. Third, the number of communication rounds. The second-order server update reaches near-peak accuracy within one to two rounds on several datasets (Figure~\ref{fig:acc_vs_rounds}), reducing the total round budget relative to first-order schemes that need many more rounds to converge.

It is important to be clear about which part of the communication cost sparsity actually reduces, and which part it leaves untouched. The savings apply to the server-to-client broadcast and to the stored model, both of which carry the readout. They do not apply to the dominant per-round cost, which is the client-to-server upload. In the federated Newton method this message carries the local gradient and the per-class Gauss-Newton Hessian rather than the readout itself, and its size is therefore independent of readout sparsity. Concretely, a client uploads $N_O N_R$ gradient entries and $N_O$ symmetric
$N_R \times N_R$ blocks per round, i.e.\ $N_O N_R (N_R+3)/2$ floats against
$N_O N_R$ for a FedAvg weight upload, a ratio of $(N_R+3)/2 \approx 250$ at
$N_R = 500$ that is independent of the dataset ($0.96$\,MB per client and round
on the binary benchmarks and $9.6$\,MB on Character, in single precision,
against $3.9$ and $39.1$\,KB). A first-order scheme would thus have to require
some $250$ times as many rounds before the cumulative uplink favoured the
second-order update, which the convex readout problem studied here does not
demand; we therefore claim efficiency for the stored model, the inference path
and the broadcast, where the sparsity of Table~\ref{tab:fl_results} gives a
$2$--$10\times$ reduction, and explicitly not for the upload. The reduction factor is quoted for the CSR payload rather than for the count of surviving weights: at $75$--$95\%$ sparsity only $5$--$25\%$ of the entries remain, which would be a $4$--$20\times$ saving on the values alone, but CSR stores a column index alongside every value, so at equal word width the realized factor is roughly halved. The design therefore trades a larger per-round upload for substantially fewer rounds and a sparse final model. Reducing the curvature payload, for example via low-rank or sketched Hessian approximations or by re-using the fixed-reservoir structure of $\mathbf{H}_k$, is a concrete and promising direction, and connects our approach to communication-efficient second-order federated methods such as FedNS~\citep{Li2024FedNS}.

\subsection{Second-Order Optimization Compatibility}
A key advantage of the twice-differentiable Smooth-L1 penalty is that the
second-order optimization it enables is exploited not only in the
centralized setting but also within the federated protocol, directly
addressing the concern that the differentiability advantage might go unused
in distributed training. Concretely, each client evaluates the local
gradient and the per-class Gauss-Newton Hessian of its Smooth-L1 loss at the
current global readout; the server sums these quantities across clients,
solves a damped Newton equation per output class to obtain the descent
direction, and selects the step size by Armijo backtracking line search
before broadcasting the updated readout. This federated Newton scheme is
made possible precisely by the smoothness of the penalty (a non-smooth L1
objective exposes no usable Hessian), and it inherits the fast, low-variance
convergence observed centrally, reaching near-peak accuracy within one to
two communication rounds on several datasets. The principal cost of
transmitting curvature information, and its interaction with the sparse-readout
communication savings, is discussed in Section~\ref{sec:comm} and is closely
related to second-order federated methods such as FedNS~\citep{Li2024FedNS}. Sharing curvature also changes what a client exposes. The transmitted block
$\mathbf{H}_{k,c} = \mathbf{X}_k^{\top}\mathrm{diag}(p_c(1-p_c))\mathbf{X}_k$ is
a label-weighted second moment of the local reservoir states rather than any raw
sequence, and is the same class of quantity that federated reservoir computing
already exchanges, the sufficient statistics of \cite{bacciu2021federated} being
its unweighted counterpart; it is nonetheless a larger exposure than a gradient
alone, since its rank is at most the local sample count, so a client holding
fewer samples than $N_R$ reveals the subspace spanned by its own states, which
the server can interpret because it holds $\mathbf{W}_{\text{in}}$ and
$\mathbf{W}_{\text{res}}$. First-order protocols are not a safe reference point
either, shared gradients being invertible to a degree in
practice~\citep{geiping2020inverting}, so the accurate statement is that the two
expose different surfaces and that ours is the larger. Because the server needs
only the sums $\sum_k \mathbf{g}_k$ and $\sum_k \mathbf{H}_k$, which are linear
in the client contributions, secure aggregation~\citep{bonawitz2017practical}
composes with this update without modifying it (Section~\ref{sec:limitations}).

\subsection{Limitations}\label{sec:limitations}

Four limitations warrant future investigation.
(1) Support identifiability: we report the \emph{level} of sparsity the readout
reaches, not the identity of the surviving reservoir coordinates. Because the
reservoir features are strongly correlated by construction
(Section~\ref{sec2}), many supports realize an equally sparse and accurate
readout, so we claim compression rather than feature selection; whether a
stable, interpretable support can be extracted is left to future work.
(2) Non-IID data: heterogeneous client distributions are not implemented in this
work, since all federated experiments use a uniform (IID) partition, and we do
not claim that the reported behavior transfers to the non-IID case. Two risks
are specific to our scheme: under client drift~\citep{shi2022optimization} the
local Gauss-Newton Hessians describe curvature around different local optima, so
their plain sum can bias the global Newton direction, and a single global
shrinkage threshold could prune directions that only minority clients rely on.
Newton-type federated optimization under statistical heterogeneity is an active
line of work~\citep{li2019feddane,safaryan2022fednl,elgabli2022fednew}, and
because our clients already expose exactly the local gradient and Gauss-Newton
Hessian these methods operate on, they are natural drop-in replacements for the
aggregation step of Algorithm~\ref{alg:dist_smoothl1_esn}; how the
$\alpha$-continuation and the shared threshold interact with
client-heterogeneous sparsity patterns is the question specific to FedSL1ESN.
(3)
Experimental scale: the main-text federated setup uses five clients
and $50$ rounds; a client-count sweep up to $K=50$
(Appendix~\ref{app2}) shows that accuracy and readout sparsity
are stable under IID partitioning, but larger client populations and
system-level effects (stragglers, partial participation) remain untested. (4) Privacy: the curvature upload exposes more than a first-order message, and
although secure aggregation applies to it unchanged, neither that nor a
differentially private variant is implemented here.

\section{Conclusion}\label{sec7}

This paper introduced SL1ESN, an Echo State Network architecture that applies Smooth-L1 regularization to the readout layer. By leveraging a twice-differentiable approximation of the L1 norm, the method keeps the readout objective amenable to second-order optimization while still driving redundant readout weights to zero. On classification, and under a single shrinkage threshold shared by every dataset and both settings, it learned readouts that are $77$--$95\%$ sparse centrally and $75$--$95\%$ sparse in the federated setting, at an accuracy close to that of the dense unregularized and L2 baselines. In the federated setting, a second-order (Newton) server update made possible by the smooth penalty converges in only a few communication rounds, and the resulting sparse readout reduces model storage, inference cost, and the server-to-client broadcast, addressing key bottlenecks for on-device deployment. The claims of this paper are confined to classification; a supplementary evaluation on two chaotic regression benchmarks (Appendix~\ref{app:regression}) indicates that the same readout tolerates far less pruning there before the autonomous rollout degrades, which delimits where the method is expected to pay off. These findings establish a foundation for sparse and efficient reservoir computing in resource-constrained environments. Future work will address adaptive regularization strategies, robustness to non-IID data distributions, for which Newton-type federated methods designed for heterogeneous data offer a concrete starting point (Section~\ref{sec:limitations}), and extensions to other reservoir computing architectures.

\section*{Code Availability}

The code implementation and extended results are available at \url{https://github.com/oroboro235/FedESN-SmoothL1}.

\section*{Acknowledgments}
This work was supported in part by the European Regional Development Fund (ERDF) under grant FKZ 2404-003-1.2 (EU-EFRE GREEN-INNO), the ProPere THWS program, and the Cybersecurity Center TTZ-WUE.

\bibliographystyle{plainnat}
\bibliography{references}

\appendix

\section{Notation}\label{app0}
\begin{table}[ht]
\centering
\footnotesize
\setlength{\tabcolsep}{4pt}
\caption{Notation summary.}
\label{tab:notation}
\begin{tabular}{@{} c p{0.62\columnwidth} @{}}
\toprule
\textbf{Symbol} & \textbf{Description} \\
\midrule
$N_I, N_R, N_O$ & Input, reservoir, and output dimensions \\
$t,\; T$ & Discrete time step; total training steps \\
$\mathbf{u}(t), \mathbf{x}(t)$ & Input vector and reservoir state \\
$\tilde{\mathbf{x}}(t)$ & Candidate (pre-leak) reservoir state \\
$\mathbf{X}$ & Matrix of collected reservoir states \\
$\hat{\mathbf{y}}(t), \mathbf{y}(t)$ & Predicted and
  target output vectors \\
$\hat{y}_c(t), y_c(t)$ & Predicted and target probability of class $c$ \\
$\mathbf{W}_{\text{in}}, \mathbf{W}_{\text{res}}, \mathbf{W}_{\text{out}}$
  & Input, reservoir, and readout weight matrices \\
$w_{ij}$ & Element of $\mathbf{W}_{\text{out}}$ \\
$\gamma$ & Leaking rate, $\gamma \in (0, 1]$ \\
$\phi, \rho$ & Connectivity and spectral radius of $\mathbf{W}_{\text{res}}$ \\
$a_{\text{in}}$ & Input scaling factor \\
$\lambda$ & Regularization strength \\
$\alpha, \alpha_0, \kappa$ & Smooth-L1 smoothing parameter, initial value,
  and factor \\
$|w|_\alpha$ & Smooth approximation of $|w|$ \\
$\mathcal{L}$ & Regularized loss function \\
$\mathbf{g}, \mathbf{H}$ & Gradient and Hessian of $\mathcal{L}$ \\
$\Delta \mathbf{w}, \eta$ & Newton update direction and (line-search) step size \\
$\mathbf{D}, \mathbf{d}_c$ & Server Newton direction matrix and its row for class $c$ \\
$\mu$ & Damping (ridge) added to the Hessian in the Newton solve \\
$\tau$ & Shrinkage threshold for readout sparsification \\
$\mathcal{S}_\tau$ & Soft-threshold operator $\mathrm{sign}(w)\max(|w|-\tau,0)$ \\
$P,\ \varepsilon_s$ & Support-stagnation early-stop window and tolerance \\
$K, k, R$ & Number of clients, client index, communication rounds\\
$\mathbf{g}_k, \mathbf{H}_k$ & Local gradient and Gauss-Newton Hessian of client $k$ \\
$\mathcal{L}_k$ & Local loss of client $k$ \\
\bottomrule
\end{tabular}
\end{table}

\section{Implementation Details}\label{app1}

\subsection{Hyperparameter Settings}

Table~\ref{tab:best_hparams} reports the per-dataset values of the spectral radius $\rho$, input scaling $a_{\text{in}}$, leaking rate $\gamma$, and regularization strength $\lambda$ selected by random search over the distributions of Table~\ref{tab:hyperparameters} for SL1ESN. The continuation schedule ($\alpha_0 = 1$, $\kappa = 2$, $\alpha_{\max} = 5\times10^{6}$) and the early-stop parameters ($P = 8$, $\varepsilon_s = 0.2$ percentage points) are fixed uniformly across all datasets. The shrinkage threshold is likewise fixed: $\tau = 0.1$ on every classification dataset, in both the centralized and the federated setting, so no per-dataset threshold is tuned anywhere in the paper. For regression, where thresholding costs long-horizon accuracy (Appendix~\ref{app:regression}), it is left at the negligible value $\tau = 10^{-5}$, so the reported regression sparsity is produced by the Smooth-L1 penalty alone. The same reservoir seed is used for all regularization methods to ensure a fair comparison.

\paragraph{Search protocol.} Model selection uses a stratified $80/20$ split of
each training set into a fitting and a validation part, with every class present
in both; the test set of Table~\ref{tab:dataset_classification} is untouched
until the final evaluation. Each candidate configuration is an independent
uniform draw from the distributions of Table~\ref{tab:hyperparameters}, scored
by validation accuracy alone, and the same budget is given to all three
penalties on a given dataset: $5{,}000$ draws on UCI HAR, Character and
JapaneseVowels and $25{,}000$ on the four binary UCR datasets. In the federated
setting the reservoir hyperparameters ($\rho$, $a_{\text{in}}$, $\gamma$) are
inherited from the centralized selection and only $\lambda$ is re-drawn, for
$100$ configurations per dataset and penalty. Each candidate is trained for up
to $100$ rounds with early stopping (patience $10$) and scored by the accuracy
of the aggregated global readout on a validation set pooled from $20\%$ of each
client's local training shard, with the threshold at the negligible value
$10^{-5}$. The selected
federated values are $\lambda = 1.1\times10^{-3}$ (UCI HAR), $1.1\times10^{-3}$
(Character), $9.7\times10^{-3}$ (JapaneseVowels), $3.4\times10^{-4}$ (ECG5000),
$3.3\times10^{-3}$ (Distal.), $1.2\times10^{-4}$ (Yoga) and $7.4\times10^{-4}$
(Strawberry). The selected configurations are then evaluated at the fixed
budget of $50$ rounds over seeds $92$--$101$ (Table~\ref{tab:fl_results}).
In the first version of this paper, the federated Strawberry value of $\lambda$ had
been carried over from an earlier search that scored candidates on test
accuracy; it has been replaced by the validation-selected value above, and the
federated Strawberry results have been recomputed accordingly.

\paragraph{Solver settings.} The damping $\mu$ of the Newton system
$(\mathbf{H} + \mu\mathbf{I})\Delta\mathbf{w} = -\mathbf{g}$ is not a fixed
constant but is set adaptively per solve, in the manner of
Levenberg--Marquardt: starting from $\mu = 1$ it is multiplied by $10$ until the
system is solvable and the resulting step is bounded
($\|\Delta\mathbf{w}\|_\infty < 10^{2}$), which is also what keeps the singular
directions of the shift-invariant softmax parameterization (Section~\ref{sec5})
from producing an unusable step. The Armijo line search uses a sufficient-decrease
constant $c_1 = 10^{-4}$ and a backtracking factor of $0.5$, capped at $20$
backtracks. The centralized iteration budget of Algorithm~\ref{alg:centralized}
is $200$ Newton iterations, which the support-stagnation criterion
($P = 8$, $\varepsilon_s = 0.2$ percentage points) reaches well before
exhaustion; the federated budget is the $50$ communication rounds of
Section~\ref{sec:dist_config}.

\begin{table}[ht]
\centering
\footnotesize
\caption{Selected SL1ESN hyperparameters per dataset (random search
over the ranges in Table~\ref{tab:hyperparameters}; see
Section~\ref{sec5}). All runs use the reservoir size $N_R = 500$.}
\label{tab:best_hparams}
\begin{tabular}{@{} l c c c c @{}}
\toprule
\textbf{Dataset} & $\rho$ & $a_{\text{in}}$ & $\gamma$ & $\lambda$ \\
\midrule
Mackey-Glass  & 1.26 & 1.76  & 0.54  & $1.5\times10^{-3}$ \\
Lorenz63      & 1.10 & 0.23  & 0.45  & $1.0\times10^{-4}$ \\
UCI HAR       & 1.48 & 7.12  & 0.016 & $6.1\times10^{-3}$ \\
Character     & 1.22 & 0.014 & 0.16  & $7.4\times10^{-3}$ \\
JapaneseVowels & 0.50 & 0.60 & 0.032 & $5.7\times10^{-4}$ \\
ECG5000       & 2.04 & 10.3  & 0.033 & $1.0\times10^{-2}$ \\
Distal.       & 1.59 & 2.58  & 0.38  & $5.2\times10^{-4}$ \\
Yoga          & 2.10 & 92.0  & 0.062 & $2.7\times10^{-3}$ \\
Strawberry    & 2.72 & 2.09  & 0.023 & $1.7\times10^{-3}$ \\
\bottomrule
\end{tabular}
\end{table}

\section{Regression on Chaotic Benchmarks}\label{app:regression}

The Smooth-L1 readout of Section~\ref{sec3} is not specific to classification, and the regression objective is trained by the same damped Newton solver with continuation and shrinkage. For completeness, this appendix evaluates the centralized SL1ESN on the Mackey-Glass~\citep{mackey1977oscillation} and Lorenz63~\citep{lorenz2017deterministic} chaotic systems, which are standard benchmarks for evaluating ESN performance on chaotic dynamics.

The setup follows the centralized protocol of Section~\ref{sec5}. Hyperparameters are optimized via random search over the distributions in Table~\ref{tab:hyperparameters}, with the valid prediction time (VPT), the number of steps for which the autonomous rollout stays within a normalized-error threshold of the true trajectory, as the accuracy objective; the selected values are listed in Table~\ref{tab:best_hparams}. In addition to the unregularized and L2 (ridge regression) readouts, we include an L1 (LASSO, coordinate descent~\citep{Wright2015}) baseline, since a direct L1 solver is available for the quadratic loss. Metrics are reported for the best of five seeds, since the autonomous rollouts of a chaotic system are too dispersed across seeds for a mean to be informative. The shrinkage threshold is left at the negligible value $\tau = 10^{-5}$, so the reported sparsity is produced by the Smooth-L1 penalty alone.

\begin{table}[ht]
\caption{Centralized regression results (best of five runs). RMSE lower is
better; VPT (valid prediction time, in steps) higher is better; sparsity is
the fraction of zero readout weights. All readouts are selected for accuracy.
\textbf{Bold}: best RMSE per dataset.}
\label{tab:regression_results}
\centering
\begin{tabular}{llcccc}
\toprule
\textbf{Dataset} & \textbf{Metric} & \textbf{none} & \textbf{L2} &
  \textbf{L1} & \textbf{SL1ESN} \\
\midrule
\multirow{3}{*}{Mackey-Glass}
  & RMSE      & $\mathbf{0.085}$ & $0.108$ & $0.113$ & $0.118$ \\
  & VPT       & $511$   & $176$   & $230$   & $377$ \\
  & Spars.\ (\%) & $0$  & $0$     & $91.0$  & $46.4$ \\
\midrule
\multirow{3}{*}{Lorenz63}
  & RMSE      & $\mathbf{9.90}$ & $10.70$ & $11.74$ & $10.14$ \\
  & VPT       & $277$   & $270$   & $109$   & $148$ \\
  & Spars.\ (\%) & $0$  & $0$     & $92.9$  & $0.2$ \\
\bottomrule
\end{tabular}
\end{table}

\begin{figure*}[htbp]
\centering
\begin{subfigure}[b]{0.24\textwidth}
  \centering
  \includegraphics[width=\textwidth]{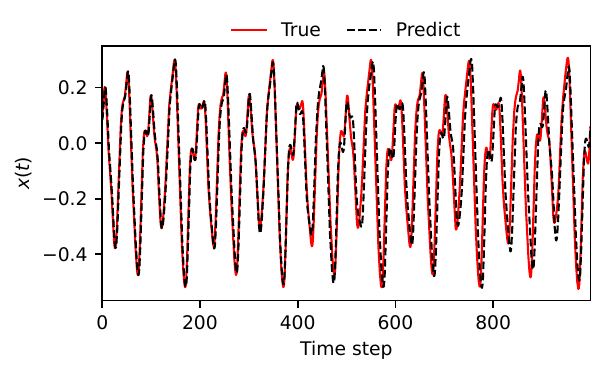}
  \caption{No regularization}
  \label{fig:mg_none}
\end{subfigure}
\hfill
\begin{subfigure}[b]{0.24\textwidth}
  \centering
  \includegraphics[width=\textwidth]{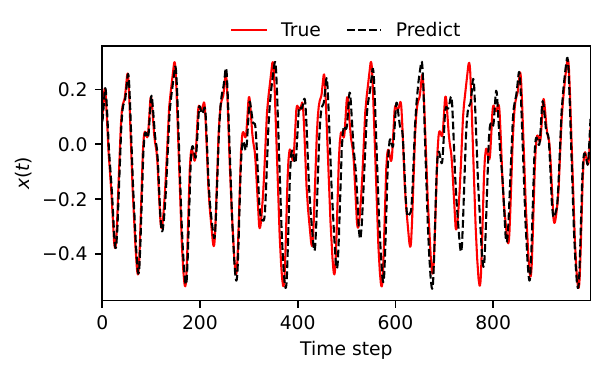}
  \caption{L2 regularization}
  \label{fig:mg_l2}
\end{subfigure}
\hfill
\begin{subfigure}[b]{0.24\textwidth}
  \centering
  \includegraphics[width=\textwidth]{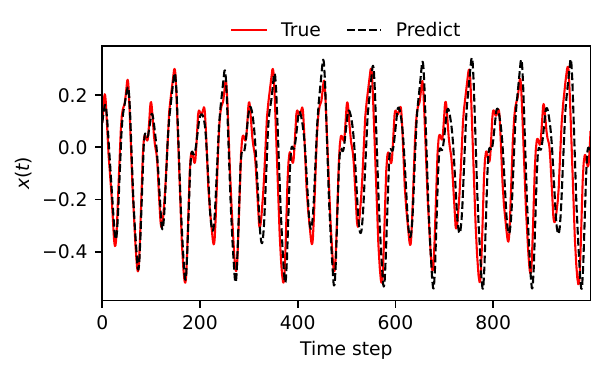}
  \caption{L1 regularization}
  \label{fig:mg_lasso}
\end{subfigure}
\hfill
\begin{subfigure}[b]{0.24\textwidth}
  \centering
  \includegraphics[width=\textwidth]{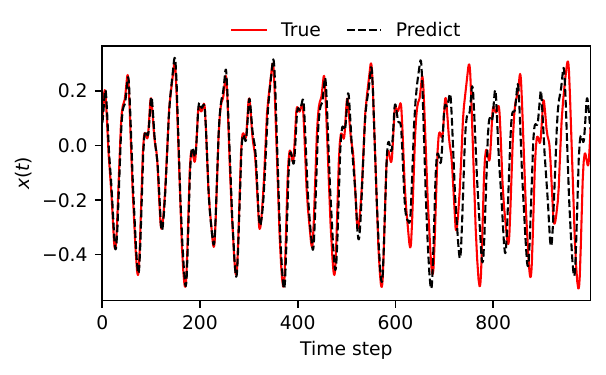}
  \caption{Smooth-L1 regularization}
  \label{fig:mg_smoothl1}
\end{subfigure}
\caption{Autonomous Mackey-Glass rollout (dashed) against the true test
trajectory (solid) under the four readout variants, at the accuracy-selected
configuration. The unregularized readout tracks the true trajectory longest
(VPT $511$; Table~\ref{tab:regression_results}), followed by the Smooth-L1
readout (VPT $377$), whereas the L2 and L1 rollouts lose phase alignment
earlier (VPT $176$ and $230$).}
\label{fig:mg_comparison}
\end{figure*}

\begin{figure*}[htbp]
\centering
\begin{subfigure}[b]{0.24\textwidth}
  \centering
  \includegraphics[width=\textwidth]{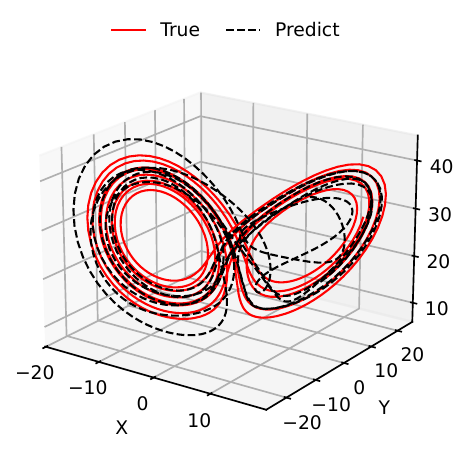}
  \caption{No regularization}
  \label{fig:lorenz_none}
\end{subfigure}
\hfill
\begin{subfigure}[b]{0.24\textwidth}
  \centering
  \includegraphics[width=\textwidth]{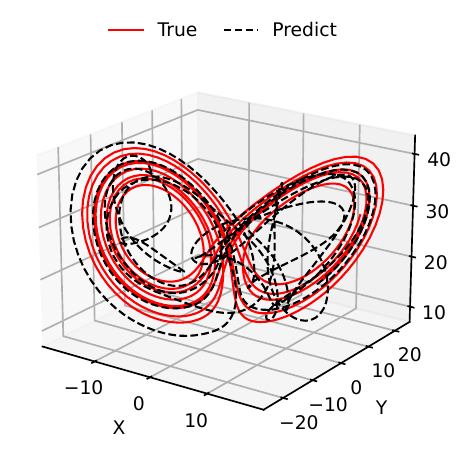}
  \caption{L2 regularization}
  \label{fig:lorenz_l2}
\end{subfigure}
\hfill
\begin{subfigure}[b]{0.24\textwidth}
  \centering
  \includegraphics[width=\textwidth]{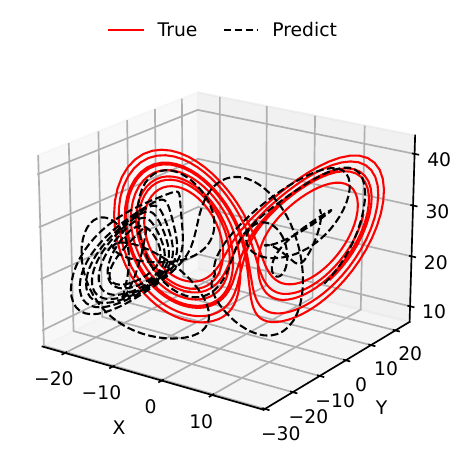}
  \caption{L1 regularization}
  \label{fig:lorenz_lasso}
\end{subfigure}
\hfill
\begin{subfigure}[b]{0.24\textwidth}
  \centering
  \includegraphics[width=\textwidth]{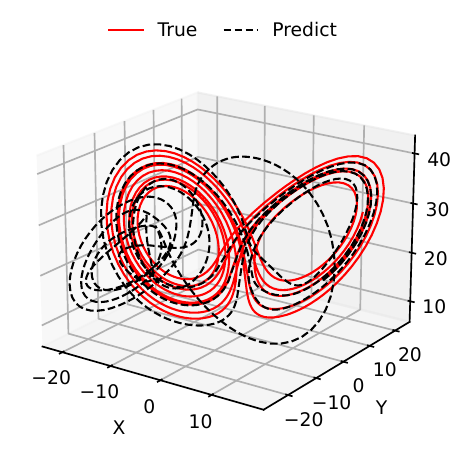}
  \caption{Smooth-L1 regularization}
  \label{fig:lorenz_smoothl1}
\end{subfigure}
\caption{Autonomous Lorenz rollout (dashed) against the true attractor
(solid) under the four readout variants, at the accuracy-selected
configuration. The unregularized, L2, and Smooth-L1 readouts all reconstruct
the butterfly and track its two lobes before slowly drifting outward (VPT
$277$, $270$, and $148$; Table~\ref{tab:regression_results}), whereas the L1
readout leaves the attractor earliest, collapsing away from the far lobe
(VPT $109$). The Smooth-L1 rollout attains the lowest RMSE of the three
regularized readouts while, unlike the coordinate-descent L1 solver,
remaining compatible with second-order optimization.}
\label{fig:regularization_comparison}
\end{figure*}

Table~\ref{tab:regression_results} reports the results, and SL1ESN is competitive on both benchmarks. On Mackey-Glass it attains RMSE $0.118$ with the second-longest valid-prediction time (VPT $377$ steps, behind only the unregularized readout), and on Lorenz it reaches the best RMSE of the regularized readouts ($10.14$), better than the L1 readout on both metrics, though its valid-prediction time (VPT $148$) trails the L2 and unregularized readouts. Figures~\ref{fig:mg_comparison} and~\ref{fig:regularization_comparison} show the corresponding rollouts. On Mackey-Glass the Smooth-L1 rollout tracks the true trajectory longest among the regularized readouts, losing phase alignment only well after the L2 and L1 rollouts. On Lorenz, the unregularized, L2, and Smooth-L1 rollouts all reconstruct the attractor and track its two lobes of the butterfly, whereas the L1 readout leaves it earliest, collapsing away from the far lobe.

The readout sparsity, however, is where the two regression tasks part company with the classification results of the main text. On Mackey-Glass the readout is $46\%$ sparse at essentially no accuracy cost, but on Lorenz the accuracy-selected readout stays effectively dense ($0.2\%$). Sparsifying it further is not free. One-step RMSE degrades gracefully, but the VPT, the horizon over which the rollout still tracks the true chaotic trajectory, is far more sensitive. Mackey-Glass can be pushed to $65\%$ sparsity at VPT $107$ and Lorenz to $36\%$ at VPT $69$, but beyond roughly $70\%$ sparsity long-horizon predictability collapses altogether, since the Mackey-Glass VPT falls below $10$ steps and the Lorenz rollout diverges almost immediately (VPT $\le 5$). The likely reason is that a chaotic attractor distributes predictive signal across many weakly-weighted, cross-dimensional reservoir couplings, so pruning by magnitude discards coordinates that the autonomous rollout still depends on, even when they contribute little to the one-step error. By contrast, the L1 (LASSO) readout does reach $91$--$93\%$ sparsity, but at a VPT well below SL1ESN on both systems, so it buys that sparsity with exactly the long-horizon fidelity at issue. We also note that, unlike in classification, where the penalty strength $\lambda$ saturates early as a sparsity control (Section~\ref{sec:sensitivity}), on these regression readouts $\lambda$ is itself an effective sparsity knob, the accuracy-selected Mackey-Glass readout already being $46\%$ sparse.

In sum, the SL1ESN regression readout is accurate but not usefully sparse, and forcing sparsity higher trades away exactly the quantity that matters for a chaotic system, namely the horizon over which the autonomous rollout remains valid. Sparse readouts, on this evidence, are a tool for classification-style tasks with redundant reservoir features rather than a universal compression method for ESNs, which is why the experimental claims of this paper are confined to classification.

\section{Sensitivity to the Number of Federated Clients}\label{app2}

\begin{figure*}[htbp]
\centering
\setlength{\abovecaptionskip}{4pt}
\includegraphics[width=0.9\textwidth]{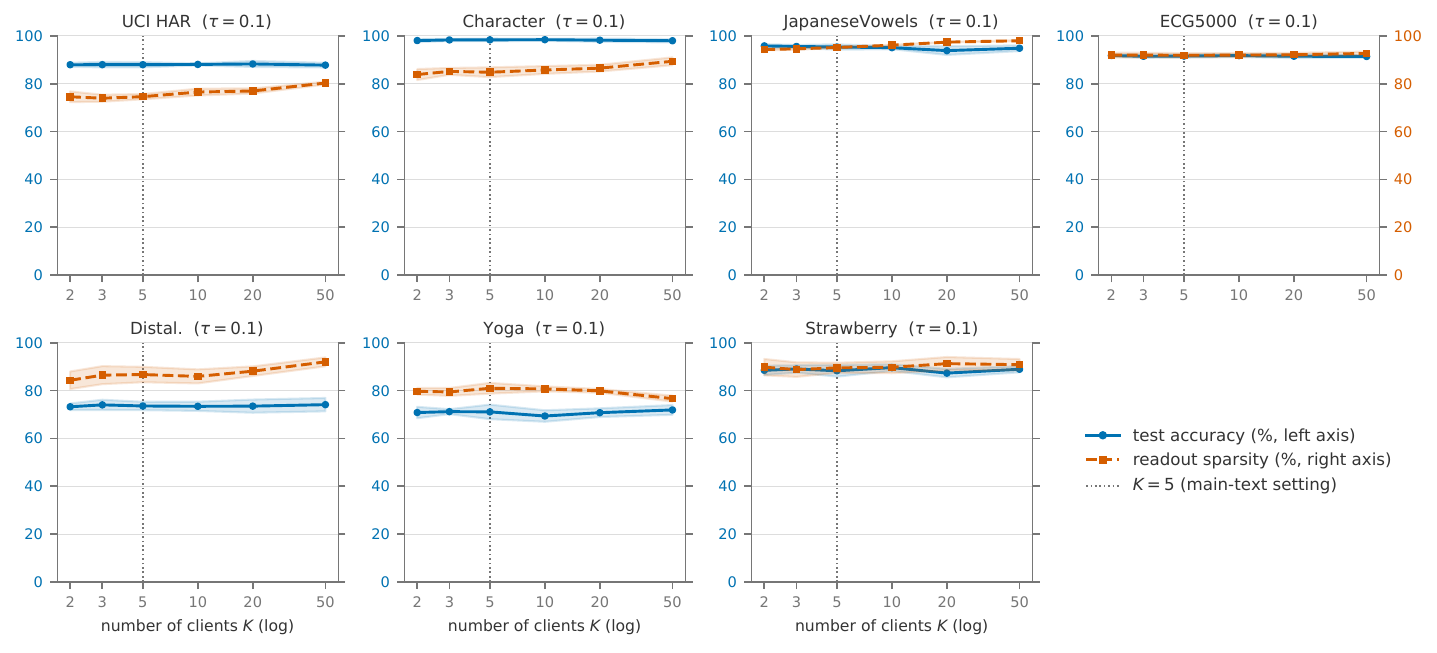}
\caption{Federated test accuracy (left axis, solid) and readout sparsity
(right axis, dashed) as the number of clients $K$ is swept from $2$ to $50$
($50$ rounds, IID partition, shared threshold $\tau = 0.1$, mean $\pm$1 std
over ten seeds). The dotted
vertical line marks the $K=5$ setting used in the main-text experiments.
Accuracy is essentially flat in $K$ on every dataset, and sparsity stays in
the band observed at $K=5$.}
\label{fig:fl_clients}
\vspace{-0.5\baselineskip}
\end{figure*}

The main-text federated experiments fix the number of clients at $K=5$ (Section~\ref{sec:dist_config}). To assess how the federated Newton scheme behaves as the data become more fragmented, we repeat the federated protocol while sweeping $K \in \{2, 3, 5, 10, 20, 50\}$ and holding every other setting fixed: $50$ communication rounds, IID partitioning, the shared threshold $\tau = 0.1$ of Table~\ref{tab:fl_results}, and ten random seeds per configuration. Figure~\ref{fig:fl_clients} reports test accuracy and readout sparsity as functions of $K$.

Two observations stand out. First, accuracy is essentially invariant to the client count. Over the entire sweep, from $K=2$ to $K=50$, the accuracy of every dataset moves within a band of at most $2.6$ points, and on four of the seven within about one point (UCI HAR $87.8$--$88.2\%$, Character $97.9$--$98.4\%$, ECG5000 $91.4$--$91.8\%$, Distal.\ $73.2$--$74.1\%$). Notably, at $K=50$ each JapaneseVowels client holds only about five training sequences, yet the global readout still reaches $94.8\%$, indistinguishable from its $K=5$ value. This robustness is expected from the structure of the update. The server operates on the sums of local gradients and Gauss-Newton Hessians, and under an IID partition the data-fidelity part of these sums is independent of how the samples are divided among clients, so $K$ enters the optimization only through the regularization term and random variation.

Second, readout sparsity also transfers across client counts, staying in the band observed at $K=5$ (e.g.\ $73.9$--$80.2\%$ on UCI HAR, $94.3$--$98.0\%$ on JapaneseVowels) with no systematic trend in $K$. This is worth stating because the aggregated objective implicitly scales the penalty with $K$ (each client's local loss carries its own copy of the Smooth-L1 term), yet no corresponding sparsity drift is visible, and fragmenting the data does not make the round-$50$ readout any less sparse. We note that this sweep retains the IID partitioning of the main text; how these conclusions change under heterogeneous client distributions remains part of the future work discussed in Section~\ref{sec:limitations}.

\end{document}